# Relational Decomposition for Program Synthesis

Céline Hocquette[1] and Andrew Cropper[2]

[1]University of Southampton
[2]University of Oxford
c.m.e.j.hocquette@soton.ac.uk; andrew.cropper@cs.ox.ac.uk

## Abstract

We introduce a relational approach to program synthesis. The key idea is to decompose synthesis tasks into simpler relational synthesis subtasks. Specifically, our representation decomposes a training input-output example into sets of input and output facts respectively. We then learn relations between the input and output facts. We demonstrate our approach using an off-the-shelf inductive logic programming (ILP) system on four challenging synthesis datasets. Our results show that (i) our representation can outperform a standard one, and (ii) an off-the-shelf ILP system with our representation can outperform domain-specific approaches.

## 1 Introduction

The goal of program synthesis is to automatically generate a computer program from a set of input-output examples [Shapiro, 1983; Gulwani *et al.*, 2017], such as a LISP [Summers, 1977], Prolog [Shapiro, 1983], or Haskell [Katayama, 2008] program. For instance, consider the examples shown in Table 1. Given these examples, we want to learn a program that inserts the letter *a* at position *2* in the input list to produce the corresponding output list.

| Input | Output |
| --- | --- |
| [l, i, o, n] | [l, a, i, o, n] |
| [t, i, g, e, r] | [t, a, i, g, e, r] |

Table 1: Input-output examples.

The standard approach to program synthesis is to search for a sequence of actions [Cropper and Dumančić, 2020; Curtis *et al.*, 2022; Aleixo and Lelis, 2023; Lei *et al.*, 2024] or functions [Lin *et al.*, 2014; Ellis *et al.*, 2018; Kim *et al.*, 2022; Ameen and Lelis, 2023; Witt *et al.*, 2025; Rule *et al.*, 2024] to map *entire* inputs to their corresponding *entire* outputs. For instance, given the examples in Table 1 and the functions *head*, *tail*, and *cons*, a system could learn the following program where $x$ is an input:

```
def f(x):
  return cons(head(x),cons('a',tail(x)))
```

Whilst the standard approach is effective for simple programs, it can struggle when learning programs that require long sequences of actions/functions. For instance, to insert the letter *a* at position *3*, a system could synthesise the program:

```
def f(x):
  return cons(head(x),cons(head(tail(x)),
         cons('a',tail(tail(x))))) 
```

This program is long and difficult to learn because the search complexity in program synthesis is exponential with the search depth [Gulwani *et al.*, 2017; Witt *et al.*, 2025]. Therefore, most existing approaches struggle to learn long sequences of actions/functions.

Rather than learn a sequence of actions/functions to map an entire input to an entire output, our key contribution is to introduce a representation that decomposes synthesis tasks into simpler relational synthesis subtasks. Specifically, our representation decomposes a training input-output example into sets of input and output facts. We then learn relations between the input and output facts.

To illustrate this idea, consider the first input-output example in Table 1. Rather than represent the example as a pair of lists, $[l,i,o,n] \mapsto [l,a,i,o,n]$, we represent the input as a set of facts of the form $in(I,V)$[1], where each fact states that the input value at index $I$ is $V$:

```
in(1,l).  in(2,i).  in(3,o).  in(4,n).
```

Similarly, rather than represent the output as a list, we represent the output as a set of facts of the form $out(I,V)$[1], where each fact states that the output value at index $I$ is $V$:

```
out(1,l). out(2,a). out(3,i). out(4,o). out(5,n).
```

We then try to generalise the *out* facts given the *in* facts and additional background knowledge, which encodes additional information about the examples. For instance, by decomposing the examples in Table 1, our approach learns the following rules as a solution:

```
out(I,V):- I<2, in(I,V).
out(2,a).
out(I,V):- I>2, in(I-1,V).
```

The first rule says that the output value at index $I$ is the input value at index $I$ for indices strictly smaller than *2*. The second rule says that the output value at index *2* is *a*. The third rule says that the output value at index *I* is the input value at index $I-1$ for indices *I* strictly greater than *2*. We can learn similar rules for *insert at position k* by learning different indices.

---
[1]We also prefix each fact with an example identifier but omit it for brevity.

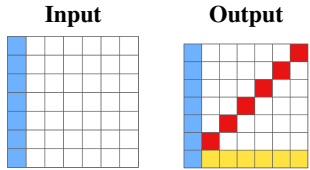

Figure 1: Input-output example for the *ARC* task *3bd67248*.

As a second illustrative scenario, consider the task shown in Figure 1, which is from the *Abstraction and Reasoning Corpus* (ARC) [Chollet, 2019]. The goal is to learn a function to map the input image to the output image. Rather than treat the input and output as entire images, we reason about individual pixels. Specifically, we represent the input image as a set of facts of the form *in(X,Y,C)*, where each fact states that the input pixel at row *X* and column *Y* has colour *C*:

```
in(1,1,blue).    in(2,1,blue).    in(3,1,blue).
in(4,1,blue).    in(5,1,blue).    in(6,1,blue).
```

We use a set of facts of the form *empty(X,Y)* to indicate that the input pixel at row *X* and column *Y* is empty/uncoloured:

```
empty(1,2).    empty(1,3).    empty(1,4).
empty(2,2).    empty(2,3).    empty(2,4).
```

Similarly, we represent the output image as a set of facts of the form *out(X,Y,C)*, where each fact states that the output pixel at row *X* and column *Y* has colour *C*:

```
out(1,1,blue).    out(7,2,yellow).    out(1,7,red).
out(2,1,blue).    out(7,3,yellow).    out(2,6,red).
```

We then try to generalise the *out* facts given the *in* and *empty* facts and additional background knowledge. For instance, we can learn the rules:

```
out(X,Y,C):- in(X,Y,C).
out(X,Y,yellow):- empty(X,Y), height(X).
out(X,Y,red):- empty(X,Y), height(X+Y-1).
```

The first rule says that any coloured pixel in the input image is the same colour in the output image. The second rule says that any uncoloured pixel in the bottom row of the input image is yellow in the output image. The last rule states that any uncoloured pixel in the input image is red in the output image if its coordinates *X* and *Y* sum to $H+1$, where *H* is the height (number of rows) of the image, i.e. if it is located on the diagonal. In other words, our representation concisely expresses the concept of a line without being given the definition.

Our representation offers several benefits. Foremost, it decomposes a task into smaller ones by decomposing each training example into multiple examples. Therefore, instead of learning a program to map an entire input list or image at once, we learn a set of rules, each generalising some list elements or image pixels. The key benefit is that we can learn each rule independently and then combine them [Cropper and Hocquette, 2023]. For instance, solving the list function task *insert at position 3* with a program that processes entire examples requires at least 8 sequential actions. In contrast, our approach only needs 3 rules, each with at most 3 literals. Since each rule is smaller, the search space is reduced, making the overall program easier to learn. The Blumer bound [Blumer *et al.*, 1987] explains why searching smaller spaces leads to better generalisation. This result says that given two search spaces, searching the smaller one is more likely to produce higher accuracy, assuming that a good program is in both.

To demonstrate our idea, we use inductive logic programming (ILP) [Muggleton, 1991; Cropper and Dumančić, 2022]. Given background knowledge and examples, the goal of ILP is to find a program that generalises the examples with respect to the background knowledge. ILP represents data and learned programs as logic programs and is therefore a relational approach to program synthesis.

**Contributions.** Our main contribution is to show that program synthesis tasks can be solved more easily if decomposed into relational learning tasks. The second contribution is to show that an off-the-shelf ILP system with our representation and a domain-independent bias can achieve high performance compared to domain-specific approaches on four varied and challenging datasets.

Overall, we make the following contributions:

- We introduce a relational representation that decomposes a synthesis task into multiple relational subtasks.
- We evaluate our representation using an off-the-shelf ILP system on four challenging datasets, including image reasoning, string transformations, and list functions. Our empirical results show that (i) our relational representation can drastically improve learning performance compared to a standard state/functional representation, and (ii) an off-the-shelf ILP system with our representation can outperform domain-specific approaches.

## 2 Related Work

**Program synthesis.** Deductive program synthesis approaches [Manna and Waldinger, 1980] deduce programs that exactly satisfy a complete specification. By contrast, we focus on *inductive program synthesis*, which uses partial specifications, typically input-output examples [Shapiro, 1983; Gulwani *et al.*, 2017]. Hereafter, *program synthesis* refers to the inductive approach. While most approaches learn functional programs [Ellis *et al.*, 2019; Shi *et al.*, 2022; Witt *et al.*, 2025; Rule *et al.*, 2024], we learn relational (logic) programs.

**Domain specific.** There are many domain-specific approaches to program synthesis, including for strings [Gulwani, 2011], 3D shapes [Tian *et al.*, 2019], list functions [Rule, 2020], and visual reasoning [Wind, 2022; Xu *et al.*, 2023; Lei *et al.*, 2024]. For instance, ICECUBER [Wind, 2022] is a symbolic synthesis approach for *ARC*. It uses 142 hand-crafted functions designed by manually solving the first 100 tasks, achieving a performance of 47%. By contrast, our approach is versatile, generalises to multiple domains, and uses an off-the-shelf general-purpose ILP system.

**State-based synthesis.** Most synthesis approaches learn a sequence of actions or functions to transform an input state to an output state. Some approaches evaluate the distance to the desired output [Ellis *et al.*, 2019; Cropper and Dumančić, 2020; Ameen and Lelis, 2023]. By contrast, we decompose examples and reason about elements or pixels.

**LLMs.** Directly comparing symbolic program synthesis to large language models (LLMs) is difficult. As Wang *et al.* [2024] state, LLMs need large pretraining datasets, which may include test data. For instance, LLMs approaches for *ARC* use datasets such as ARC-Heavy (200k tasks) or ARC-Potpourri (400k tasks) [Li *et al.*, 2024], additional training examples [Hodel, 2024], or data augmentations [Franzen *et al.*, 2024]. The ARChitects [Franzen *et al.*, 2024], winners of the ARC-AGI challenge, pretrained their solution on 531,318 examples. By contrast, our approach requires no pretraining and uses only the 2-10 training examples provided for each task.

**Lists and images.** Our experiments focus on synthesis tasks over lists and images. Lists are a simple yet expressive domain, well-suited to representing observations in many domains such as computational biology, where proteins, genes, and DNA are typically represented as strings [Raedt, 2008]. As Rule [2020] explains, lists use numbers in multiple roles (symbols, ordinals, and cardinals) and support recursive structures. Lists naturally align with familiar psychological concepts, encompass classic concept learning domains, and are formally tractable. Similarly, image tasks, like those in the ARC [Chollet, 2019], capture a wide range of abstract concepts, including shapes, patterns, and spatial relationships. They offer high task diversity and align well with human core knowledge priors.

**ILP.** Many ILP approaches use state representations [Lin *et al.*, 2014; Cropper and Dumančić, 2020]. Related approaches with decomposed representations include Silver *et al.* [2020] and Evans *et al.* [2021]. These approaches are specifically designed for learning game policies from demonstrations and dynamics from temporal sequences, respectively. By contrast, we use a general-purpose off-the-shelf ILP system and consider program synthesis tasks.

**Decomposition.** Some approaches partition the training examples into subsets, learn programs for each subset, and combine them into a global solution [Cropper and Hocquette, 2023]. By contrast, we decompose each training example into multiple examples. BEN [Witt *et al.*, 2025] decomposes examples into objects, aligns them through analogical reasoning, and synthesises programs for the resulting subtasks. We differ in many ways. First, while BEN uses domain-specific rules to decompose an example into objects, we simply decompose lists and images into individual elements. Second, BEN relies on hand-engineered functions, such as *border(s)*, which draws a border of size *s* and *denoise(s)*, which denoises an object in the *ARC* domain, whereas we use only basic arithmetic operations like addition. Finally, BEN synthesises functional programs that manipulate object-based states, whereas we learn relational rules between input and output elements.

**Representation change.** Representation change refers to changing the language used to represent knowledge, including the examples [Cohen, 1990]. Bundy [2013] argues that finding the right representation is the key to successful reasoning. We contribute to this view by showing that simply looking at a problem differently can greatly improve learning performance.

## 3 Problem Setting

We formulate the synthesis problem as an ILP problem. We assume familiarity with logic programming [Lloyd, 2012] but provide a summary in the appendix. We use the term *rule* synonymously with *definite clause*. A *definite program* is a set of definite clauses with the least Herbrand model semantics. We refer to a definite program as a *logic program*. A *hypothesis space* is a set of hypotheses (logic programs) defined by a language bias, which restricts the syntactic form of hypotheses [Cropper and Dumančić, 2022].

We use the learning from entailment setting of ILP [Raedt, 2008]. We define an ILP task:

**Definition 1** (ILP task). An ILP task is a tuple $(E^+, E^-, B, \mathcal{H}, cost_{B,E^+,E^-})$, where $E^+$ and $E^-$ are sets of ground atoms denoting positive and negative examples respectively, $B$ is a logic program denoting the background knowledge, $\mathcal{H}$ is a hypothesis space, and $cost_{B,E^+,E^-} : \mathcal{H} \mapsto \mathbb{N}$ is a function that measures the cost of a hypothesis.

We define an optimal hypothesis:

**Definition 2** (Optimal hypothesis). For an ILP task $(E^+, E^-, B, \mathcal{H}, cost_{B,E^+,E^-})$, a hypothesis $h \in \mathcal{H}$ is optimal when $\forall h' \in \mathcal{H}, cost_{B,E^+,E^-}(h) \leq cost_{B,E^+,E^-}(h')$.

In this paper, we assume a noiseless setting. We search for a hypothesis $h$ which entails all examples in $E^+$ ($\forall e \in E^+, h \cup B \models e$) and no example in $E^-$ ($\forall e \in E^-, h \cup B \not\models e$). A hypothesis has an infinite cost if it does not entail all positive examples or if it entails any negative examples. Otherwise, its cost is equal to its size (number of literals in the hypothesis).

## 4 Decomposing Examples

Rather than learn a sequence of actions/functions to map an entire input to an entire output, we introduce a representation that decomposes synthesis tasks into simpler relational subtasks. Specifically, our representation decomposes a training input-output example into input and output facts.

Algorithm 1 shows our algorithm for decomposing examples. It takes as input a set of examples $E$ and a domain $D$ of element values. $E$ is a set of input-output examples of the form $i \mapsto o$, where $i$ is an $n$-dimensional array and $o$ is an $m$-dimensional array. Algorithm 1 returns a tuple $(E^+, E^-, B)$. We consider each example $i \mapsto o$ in turn, and define an identifier *id* for the current example (line 5). For each element $x$ in $i$, we generate the fact *in(id,$I_1$,...,$I_n$,V)*, where $(I_1,...,I_n)$ is the position of $x$ in $i$ and $V$ its value. We add this fact to the background knowledge $B$ (line 9). For each element $y$ in $o$, we generate the fact *out(id,$I_1$,...,$I_m$,V)*, where $(I_1,...,I_m)$ is the position of $y$ in $o$ and $V$ its value. We add this fact to the positive examples $E^+$ (line 13). We reason under the closed-world assumption [Reiter, 1977] to generate negative examples. For each element $y$ in $o$ and for each value $W$ in the domain of $V$, where $W \neq V$, we generate the negative example *out(id,$I_1$,...,$I_m$,W)*, where $(I_1,...,I_m)$ is the position of $y$ in $o$ and $V$ its value. We add this fact to the negative examples $E^-$ (line 16).

In the next section, we empirically show that using a decomposed representation can substantially improve learning performance.

**Algorithm 1** Example Decomposition

```
1 def decompose(E, D):
2   E⁺, E⁻, B = {}, {}, {}
3   id = 0
4   for i ↦ o in E:
5     id += 1
6     for x in i:
7       let (I₁,...,Iₙ) be the position of x in i
8       let V be the value of x in i
9       B += in(id,I₁,...,Iₙ,V)
10    for y in o:
11      let (I₁,...,Iₘ) be the position of y in o
12      let V be the value of y in o
13      E⁺ += out(id,I₁,...,Iₘ,V)
14      for W in D:
15        if W ≠ V:
16          E⁻ += out(id,I₁,...,Iₘ,W)
17  return E⁺, E⁻, B
```

## 5 Evaluation

To test our claim that decomposing a synthesis task can improve learning performance, our evaluation aims to answer the question:

**Q1** Can our decomposed representation outperform a standard state/functional representation?

To answer **Q1**, we compare the learning performance of an ILP system with a decomposed representation (Algorithm 1) against a state/functional representation. We use the same ILP system so the only difference is the representation.

To test our claim that our decomposed representation is competitive with domain-specific approaches, our evaluation aims to answer the question:

**Q2** Can a general-purpose ILP system with a decomposed representation outperform domain-specific approaches?

To answer **Q2**, we compare the learning performance of a general-purpose ILP system with a decomposed representation against domain-specific approaches.

### 5.1 Datasets

We use the following diverse and challenging datasets.

**1D-ARC.** The *1D-ARC* dataset [Xu *et al.*, 2024] is a one-dimensional adaptation of *ARC*.

**ARC.** The *ARC* dataset [Chollet, 2019] evaluates to perform abstract reasoning and problem-solving from a small number of examples. The goal of each task is to transform two-dimensional input images into their corresponding output images. The tasks are widely varied, including pattern recognition, geometric transformations, colour manipulation, and counting. We use the *training* subset and report top-1 accuracy, following related work [Witt *et al.*, 2025; Xu *et al.*, 2023; Xu *et al.*, 2024; Wang *et al.*, 2024].

**Strings.** The goal is to learn string transformation programs [Lin *et al.*, 2014]. This real-world dataset gathers user-provided examples from online forums and is inspired by a dataset of user-provided examples in Microsoft Excel [Gulwani, 2011].

**List functions.** This dataset [Rule, 2020; Rule *et al.*, 2024] evaluates human and machine learning ability. The goal of each task is to identify a function that maps input lists to output lists, where list elements are natural numbers. The tasks range from basic list functions, such as duplication and removal, to more complex functions involving conditional logic, arithmetic, and pattern-based reasoning.

### 5.2 Decomposed Representation

We use a purposely simple bias formed of the decomposed training examples and basic relations for arithmetic addition and value comparison. We describe our bias for each domain.

**1D-ARC.** We decompose a one-dimensional image into a set of pixel facts. The fact *empty(I)* holds if the pixel at index *I* is a background pixel (an uncoloured pixel). We allow integers between 0 and 9, representing the 10 different colours, as constant symbols.

**ARC.** We decompose a two-dimensional image into a set of pixel facts. The fact *empty(X,Y)* holds if the pixel at row *X* and column *Y* is a background pixel. We allow integers between 0 and 9 as constant symbols. We use the relations *height* and *width* to identify the dimensions of the image, *midrow* and *midcol* to locate the middle row and column, respectively, and *different* to determine colour inequality.

**Strings.** We decompose a string into a set of character facts. The fact *end(I)* denotes the end position of an input string. We use the relation *changecase* to convert a lowercase letter to uppercase or vice versa.

**List functions.** We decompose a list into a set of element facts. The fact *end(I)* denotes the end position of an input list. Following Rule [2020], we allow integers between 0 and 9 for the first 80 problems and integers between 0 and 99 for the remaining ones.

### 5.3 Existing Representations

We compare our approach against three standard (undecomposed) representations from the literature.

**Undecomposed list (UD-List).** We use a functional representation designed for list functions tasks [Rule, 2020] which contains the relations *head*, *tail*, *empty*, and *cons*.

**Undecomposed element (UD-Elem).** We extend the *UD-List* representation with the relations *element_at* and *empty_at* to extract elements/pixels in lists/images.

**Undecomposed string (UD-Str).** We use a functional representation designed for string transformation tasks [Lin *et al.*, 2014] which recursively parses strings left to right.

We also use the same arithmetic relations and constant symbols as in the decomposed representation for each domain. Although we aim to provide similar relations for all representations, the biases in these undecomposed representations differ from those in the decomposed representation.

### 5.4 Systems

We use the following systems.

**POPPER.** We use the ILP system POPPER [Cropper and Morel, 2021] because it can learn large programs, especially programs with many independent rules [Cropper and Hocquette, 2023].

**ARGA.** ARGA [Xu *et al.*, 2023] is an object-centric approach designed for *ARC*. ARGA abstracts images into graphs and then searches for a program using a domain-specific language. ARGA uses 15 operators, such as to rotate, mirror, fill, or hollow objects.

**METABIAS (MB).** The ILP system METABIAS [Lin *et al.*, 2014] uses a functional representation specifically designed for the string transformations dataset that we consider in our experiments. It uses 11 operators, such as to copy a word and convert a word to uppercase or lowercase.

**BEN.** BEN [Witt *et al.*, 2025] decomposes images into objects and learns a functional program. It uses 15 object features and 11 relations for *ARC*, and 14 features and 11 relations for *strings*[2]. See Section 2 for more details on BEN.

**HL.** Hacker-Like (HL) [Rule, 2020; Rule *et al.*, 2024] is an inductive learning system designed for the *list functions* dataset and using Monte Carlo tree search. HL aims to reproduce human learning, rather than outperform it. However, it outperforms other program synthesis approaches on the *list functions* dataset such as METAGOL [Muggleton *et al.*, 2015], ROBUSTFILL[3] [Devlin *et al.*, 2017], CODEX [Chen *et al.*, 2021], and FLEET [Yang and Piantadosi, 2022]. Among these, only HL and FLEET achieve human-level performance, while the others greatly struggle.

### 5.5 Experimental Setup

We measure predictive accuracy (the proportion of correct predictions on test data). For our decomposed representation, a prediction is correct only if all output elements/characters/pixels are correct. For the *strings* and *list functions* datasets, we perform leave-one-out cross-validation. For tasks 81 to 250 in the *list functions* dataset, due to the large number of constant values, we sample 10,000 negative examples per task. We repeat each learning task 3 times and calculate the mean and standard error. The error values in the tables represent the standard error. We use an Intel compute node with dual 2.0 GHz Intel Xeon Gold 6138 processors, 40 CPU cores, and 192 GB of DDR4 memory. Each system uses a single CPU. We describe our experimental setup for each research question.

**Q1.** We compare POPPER with our decomposed representation against POPPER with undecomposed representations.

**Q2.** We compare POPPER with our decomposed representation against domain-specific approaches (ARGA, METABIAS, BEN, and HL).

### 5.6 Results

#### Q1: Can our decomposed representation outperform a standard state/functional representation?

Table 2 shows the results. It shows that our decomposed representation outperforms all undecomposed ones on all four domains and for all maximum learning times except *UD-Str* on the *strings* dataset. A McNeymar's test confirms the statistical significance ($p < 0.01$) of the difference. For instance,

---

[2]The code of BEN is not publicly available, and the authors were unable to share it with us. As a result, we show the results reported in their paper. Since the evaluation was performed on different hardware, the comparison should be viewed as indicative only.

[3]ROBUSTFILL required 3 days of training which highlights the search efficiency of our approach.

POPPER with our decomposed representation achieves 71% accuracy on *strings* compared to 21% for *UD-List*.

| Dataset | Time | UD-List | UD-Elem | UD-Str | Decom |
|---|---|---|---|---|---|
| 1DARC | 1 | $0 \pm 0$ | $0 \pm 0$ | $0 \pm 0$ | $\mathbf{59 \pm 7}$ |
| | 10 | $0 \pm 0$ | $0 \pm 0$ | $0 \pm 0$ | $\mathbf{63 \pm 7}$ |
| | 60 | $0 \pm 0$ | $0 \pm 0$ | $0 \pm 0$ | $\mathbf{69 \pm 6}$ |
| *ARC* | 1 | $0 \pm 0$ | $0 \pm 0$ | $0 \pm 0$ | $\mathbf{15 \pm 1}$ |
| | 10 | $0 \pm 0$ | $0 \pm 0$ | $0 \pm 0$ | $\mathbf{20 \pm 1}$ |
| | 60 | $0 \pm 0$ | $0 \pm 0$ | $0 \pm 0$ | $\mathbf{22 \pm 1}$ |
| Strings | 1 | $15 \pm 2$ | $11 \pm 2$ | $\mathbf{55 \pm 3}$ | $54 \pm 3$ |
| | 10 | $17 \pm 2$ | $16 \pm 2$ | $\mathbf{77 \pm 2}$ | $68 \pm 2$ |
| | 60 | $21 \pm 2$ | $19 \pm 2$ | $\mathbf{79 \pm 2}$ | $71 \pm 2$ |
| Lists | 1 | $10 \pm 1$ | $8 \pm 1$ | $2 \pm 1$ | $\mathbf{27 \pm 2}$ |
| | 10 | $12 \pm 1$ | $11 \pm 1$ | $2 \pm 1$ | $\mathbf{46 \pm 2}$ |
| | 60 | $14 \pm 1$ | $13 \pm 1$ | $2 \pm 1$ | $\mathbf{52 \pm 2}$ |

Table 2: Predictive accuracy (%) of POPPER with our decomposed representation versus undecomposed representations for different maximum learning times (mins).

One reason for the performance improvement is that our representation decomposes a task into multiple subtasks. For instance, consider the *ARC* task *253bf280* shown in Figure 2. The goal is to colour in green pixels in between two blue pixels in the input image. Our approach learns the rules:

```
out(X,Y,blue):- in(X,Y,blue).
out(X,Y,green):- in(X,Y1,blue), in(X,Y2,blue), Y1<Y<Y2.
out(X,Y,green):- in(X1,Y,blue), in(X2,Y,blue), X1<X<X2.
```

The first rule says that an *out* pixel is blue if it is blue in the input. The second rule says that an *out* pixel is green if it is between two blue pixels in the same row in the input. The third rule says that an *out* pixel is green if it is between two blue pixels in the same column in the input. In other words, our approach learns one rule for the permanence of blue pixels, one for horizontal lines, and one for vertical lines. Moreover, our approach learns this perfect solution without being given the definition of a line. By contrast, POPPER cannot solve this task with any of the undecomposed representations tested.

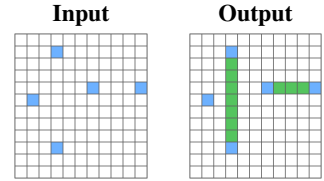

Figure 2: *ARC* task *253bf280*.

Similarly, consider the *string* task *117* shown in Table 3. The goal is to capitalise the first letter of both the first and last names. For this task, our approach learns the rules:

```
out(1,C):- in(1,C1), changecase(C1,C).
out(I,C):- in(I,C1), changecase(C1,C), in(I-1,' ').
out(I,C):- in(I,C), in(I-1,C1), changecase(C1,C2).
```

The first rule says that the *out* character at index 1 is the *in* character at index 1 uppercased. The second rule says that the *out* character at index $I$ is the *in* character at index $I$ uppercased if the *in* character at index $I-1$ is a space. The last rule says that the *out* character at index $I$ is the *in* character at index $I$ if the *in* character at index $I-1$ is a lowercase letter. In other words, our approach learns three rules: one for uppercasing the first letter of the first name, one for the last name, and one for copying the remaining lowercase letters. It learns this program without being given the definition of a word. By contrast, POPPER cannot solve this task with any of the undecomposed representations tested.

| Input | Output |
|---|---|
| joanie faas | Joanie Faas |
| oma cornelison | Oma Cornelison |

Table 3: *String* task 117.

Another reason for the performance improvement is that our decomposed representation allows programs to be expressed more compactly. In program synthesis, the search space grows exponentially with the size of the target program. By using a more compact representation, we reduce the size of the search space. The Blumer bound [Blumer *et al.*, 1987] theoretically explains why searching smaller spaces leads to better generalisation. This result says that given two search spaces, searching the smaller one is more likely to produce higher accuracy, assuming that a good program is in both. For instance, the goal of the *ARC* task *6d75e8bb* (Figure 3) is to colour in red empty pixels within the rectangle delimited by blue pixels. Our approach learns the rules:

```
out(X,Y,C):- in(X,Y,C).
out(X,Y,red):- empty(X,Y), in(X,Y1,C), in(X1,Y,C).
```

The first rule says that an *out* pixel has colour $C$ if it has colour $C$ in the input. The second rule says that an *out* pixel is red if it is empty in the input and if there are a pixel in the same row ($X$) and a pixel in the same column ($Y$) with the same colour ($C$) in the input. In other words, our approach compactly captures the concept of a rectangle without being given the definition in the background knowledge.

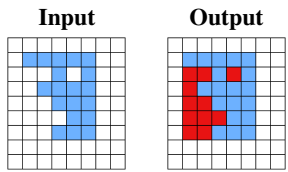

Figure 3: *ARC* task *6d75e8bb*.

POPPER struggles with our decomposed representation to solve some tasks due to our purposely simple bias. For instance, the goal of the *ARC* task *5582e5ca* is to learn a program that colours the output image with the majority colour from the input image. However, we do not include a counting mechanism in our bias and POPPER struggles to learn how to count, as it requires reasoning over all pixels.

Overall, these results suggest that the answer to **Q1** is yes: our decomposed representation can outperform an undecomposed one.

## Q2: Can a general-purpose ILP system with a decomposed representation outperform domain-specific approaches?

Table 4 shows the results. It shows that the general-purpose ILP system POPPER with our decomposed representation is competitive with, and can outperform, domain-specific approaches. Notably, it outperforms ARGA on *ARC*, METABIAS on *strings* and HL on *lists*. A McNeymar's test confirms the significance ($p < 0.01$) of these differences. We discuss the results for each dataset in turn.

| Dataset | Time | ARGA | BEN | MB | HL | Decom |
|---|---|---|---|---|---|---|
| 1DARC | 1 | **93**±**6** | na | 0±0 | 0±0 | 59±7 |
|  | 10 | **94**±**6** | na | 0±0 | 0±0 | 63±7 |
|  | 60 | **94**±**6** | na | 0±0 | 0±0 | 69±6 |
| ARC | 1 | 8±1 | 6±na | 0±0 | 0±0 | **15**±**1** |
|  | 10 | 11±2 | **25**±**na** | 0±0 | 0±0 | 20±1 |
|  | 60 | 12±2 | na | 0±0 | 0±0 | **22**±**1** |
| Strings | 1 | 0±0 | 85±na | 25±2 | 0±0 | 54±3 |
|  | 10 | 0±0 | na | 25±2 | 0±0 | **68**±**2** |
|  | 60 | 0±0 | na | 26±2 | 0±0 | **71**±**2** |
| Lists | 1 | 0±0 | na | 0±0 | **31**±**2** | 27±2 |
|  | 10 | 0±0 | na | 7±1 | 33±2 | **46**±**2** |
|  | 60 | 0±0 | na | 8±1 | 35±3 | **52**±**2** |

Table 4: Predictive accuracy (%) of our decomposed representation versus domain-specific systems for different maximum learning times (mins)[4].

**1D-ARC** ARGA outperforms our decomposed representation on the *1D-ARC* dataset (94% vs 69% predictive accuracy with a maximum learning time of 1h). This result is unsurprising because ARGA is designed for image reasoning tasks and uses domain-specific operators, such as the ability to fill, mirror, and hollow objects. This background knowledge is particularly useful for tasks such as *fill*, *mirror*, and *hollow*. By contrast, our decomposed representation is not designed for these tasks and does not include domain-specific relations.

Our decomposed representation significantly outperforms HL on the *1D-ARC* dataset (69% vs 0% predictive accuracy with a maximum learning time of 1h). Although these tasks involve identifying list functions, HL struggles on them. We asked the authors of HL for potential explanations and they explained that HL does not perform as well on problems requiring a recursive solution as it does on non-recursive problems. For instance, the task *denoise* (Figure 4) requires learning a recursive solution with an undecomposed representation, which is difficult for HL. By contrast, our approach learns the non-recursive rule:

```
out(I,C):- in(I1,C), in(I1+1,C), I2<2, I1+I2=I.
```

This rule says that an *out* pixel at index $I$ has colour $C$ if there are two adjacent pixels with colour $C$ in the input image (at indices $I1$ and $I1+1$), where one of these pixels is at index $I$ (if $I2 = 0$, then $I1 = I$, and if $I2 = 1$, then $I1 + 1 = I$), i.e. the pixel at index $I$ has an adjacent pixel with the same colour. This rule generalises perfectly to the test data. Notably, unlike ARGA and BEN, which both use a *denoise* operator, our approach learns this rule without domain-specific operators.

---

[4]We report results for 1 and 10 minutes for *ARC* and 1 min for *strings* for BEN, as they are the only ones provided in the paper.

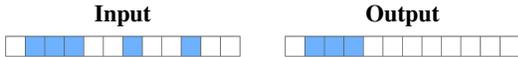

| Input | Output |

Figure 4: The *denoise* task from *1D-ARC*.

**ARC** POPPER with our decomposed representation outperforms ARGA on the *ARC* (22% vs 12% accuracy with a maximum learning time of 1h). ARGA struggles, partly because it assumes input and output images have identical sizes, preventing it from solving 138/400 tasks with different sizes.

HL assumes inputs and outputs are one-dimensional lists, so it struggles on *ARC*. METABIAS uses a bias for string manipulation, such as skipping words or converting them to uppercase, so struggles on both *1D-ARC* and *ARC*, where values are all digits.

BEN achieves 25% accuracy given a 10 mins search timeout. BEN uses a hand-designed domain-specific function to decompose images into objects. Without this function, the accuracy of BEN drops to 6%. Moreover, BEN uses hand-designed *ARC*-specific functions, such as *mirror*, *inner*, and *denoise*. By contrast, we only use general background knowledge, such as how to add two numbers.

**Strings** ARGA achieves default performance (0%) because it cannot identify any object. Similarly, HL achieves only default performance, as it assumes constants are numbers and cannot handle characters.

POPPER with our decomposed representation significantly outperforms METABIAS on the *string* dataset and achieves 71% vs 26% with a maximum learning time of 1h. The reason is that POPPER outperforms METAGOL on which METABIAS is based [Cropper and Hocquette, 2023].

BEN outperforms POPPER with our representation (85% versus 54% in 1min). BEN uses domain-specific background functions, such as capitalising the first character, dropping the first *n* characters, and adding a white space, as well as features, such as the number of digits, uppercase and lowercase letters. By contrast, we only use general background knowledge.

**List functions** POPPER with our decomposed representation significantly outperforms ARGA on the *list functions* dataset (52% vs 0% in 1h). ARGA struggles because it requires inputs and outputs of identical size, which prevents it from solving 188/250 tasks. Additionally, ARGA is designed for object-centric tasks so struggles when it cannot identify meaningful objects. Finally, ARGA uses operators designed for image reasoning, which do not generalise well to list functions. By contrast, our approach generalises to a broader range of problems, partly because we use a domain-independent bias.

POPPER with our decomposed representation significantly outperforms HL (52% vs 35% in 1h). Since HL is designed to reproduce human learning, it is understandable that our approach performs better. For instance, the goal of the *list function* task *194* (Table 5) is to reverse the input list and add its length at the start and end. Humans achieve less than 25% accuracy on this task [Rule, 2020] and HL achieves 0%. By contrast, we achieve 100% accuracy with the rules:

```
out(1,E-1):- end(E).
out(I,V):- end(E), add(I,I1,E+1), in(I1,V).
out(E+1,E-1):- end(E).
```

The first rule says that the *out* element at index 1 is $E - 1$, where $E$ is the index of the first empty position in the input list. The last rule says that the *out* element at index $E + 1$ is $E - 1$, where $E$ is the index of the first empty position in the input list. The second rule says that the *out* element at index $I$ is the *in* element at index $I1$, where $I + I1 = E + 1$. In other words, this second rule compactly expresses the concept of reverse and move by one position.

| Input | Output |
| --- | --- |
| [81, 43] | [2, 43, 81, 2] |
| [1, 63, 21, 16] | [4, 16, 21, 63, 1, 4] |

Table 5: *List function* task 194.

Overall, the results suggest that the answer to **Q2** is yes: a general-purpose ILP system with a decomposed representation can outperform domain-specific approaches.

## 6 Conclusions and Limitations

We have introduced a representation that decomposes synthesis tasks into smaller relational subtasks. Our empirical results on four domains show that our decomposed representation substantially outperforms an undecomposed one. Moreover, we show that an off-the-shelf ILP system using our decomposed representation with little domain-specific bias is competitive with, and in some cases outperforms, highly engineered domain-specific approaches. More broadly, our results show that simply representing a problem differently can greatly improve learning performance.

### 6.1 Limitations

**Bias.** In our evaluation, we use a purposely simple bias formed of raw input (elements or pixels) and basic arithmetic relations. While this simple bias achieves good performance, it is limiting for some tasks. For instance, by adding relations describing the ordinality of some characters in the *strings* domain, we improve the predictive accuracy by 9% and 4% with maximum learning times of 1 minute and 60 minutes, respectively. Future work should explore adding general-purpose concepts, such as counting, to further improve performance.

**ILP system.** We have shown that an off-the-shelf ILP system is competitive with domain-specific approaches. However, this system struggles on some tasks, where a good solution exists in the search space but the system cannot find it within the time limit. This limitation is due to the system we use and not our representation. However, our decomposed representation is system-agnostic, so we could use alternative ILP systems. Moreover, because we use an off-the-shelf ILP system, our approach naturally benefits from any developments in ILP.

## 7 Appendices, Code, and Data

The experimental code and data are available at https://github.com/celinehocquette/ijcai25-relational-decomposition.

## Acknowledgements

The authors acknowledge the use of the IRIDIS High Performance Computing Facility at the University of Southampton in completing this work.

- Section A provides background on the terminology used in the paper.
- Section C shows the programs learned by POPPER with our relational representation.

## A Terminology

### A.1 Logic Programming

We assume familiarity with logic programming [Lloyd, 2012] but restate some key relevant notation. A *variable* is a string of characters starting with an uppercase letter. A *predicate* symbol is a string of characters starting with a lowercase letter. The *arity* $n$ of a function or predicate symbol is the number of arguments it takes. A constant symbol is a function or a predicate symbol with arity zero. A variable is *first-order* if it can be bound to a constant symbol or another first-order variable. A term is a variable or a constant symbol. An *atom* is a tuple $p(t_1, ..., t_n)$, where $p$ is a predicate of arity $n$ and $t_1, ..., t_n$ are terms, either variables or constants. An atom is *ground* if it contains no variables. A *literal* is an atom or the negation of an atom. A *clause* is a set of literals. The variables in a clause are universally quantified. A *clausal theory* is a set of clauses. A *constraint* is a clause without a positive literal. A *definite clause* or *rule* is a clause with exactly one positive literal. A *definite program* is a set of definite clauses with the least Herbrand semantics. A *program* or *hypothesis* is a definite program.

## B Datasets

**1D-ARC.** The *1D-ARC* dataset contains 18 tasks, each with 3 training examples and 1 testing example. Each pixel is one of 10 different colours.

**ARC.** We use the *training* subset, which contains 400 tasks, each with 2 to 10 training examples. Image sizes range between 1x1 and 30x30 pixels. Input and output images can have different sizes. Each pixel is one of 10 different colours.

**Strings.** The dataset contains 123 tasks, each with 10 examples.

**Lists.** The dataset contains 250 tasks, each with 11 examples.

## C Programs

POPPER with our relational representation learns the following programs on the *onedarc* domain:

```
% 1d_denoising_1c
out(V0,V1,V2):- in(V0,V3,V2),add(V3,V4,V1),
                succ(V3,V5),in(V0,V5,V2),add(V1,V6,V5).

% 1d_fill
out(V0,V1,V2):- in(V0,V5,V2),add(V1,V3,V5),
                in(V0,V6,V2),add(V4,V6,V1).

% 1d_hollow
out(V0,V1,V2):- in(V0,V1,V2),succ(V4,V1),c3(V6),
                empty(V0,V3),add(V4,V5,V3),lt(V5,V6).

% 1d_move_1p
out(V0,V1,V2):- succ(V3,V1),in(V0,V3,V2).

% 1d_pcopy_1c
```

```
out(V0,V1,V2):- in(V0,V5,V2),add(V1,V6,V5),
                lt(V6,V1),succ(V3,V5),add(V3,V4,V1).
out(V0,V1,V2):- succ(V4,V1),in(V0,V4,V2),
                succ(V3,V4),empty(V0,V3).

% 1d_denoising_mc
out(V0,V1,V2):- c8(V3),in(V0,V3,V2),in(V0,V1,V4).

% 1d_mirror
out(V0,V1,V2):- v9(V3),in(V0,V4,V3),add(V4,V5,V1),
                add(V5,V6,V4),in(V0,V6,V2).

% 1d_move_2p
out(V0,V1,V2):- c2(V3),add(V3,V4,V1),in(V0,V4,V2).

% 1d_move_3p
out(V0,V1,V2):- c3(V3),add(V3,V4,V1),in(V0,V4,V2).

% 1d_pcopy_mc
out(V0,V1,V2):- c2(V4),in(V0,V3,V2),add(V4,V5,V3),
                lt(V5,V1),empty(V0,V5),add(V3,V4,V6),
                lt(V1,V6),empty(V0,V6).

% 1d_scale_dp
out(V0,V1,V2):- in(V0,V1,V2).
out(V0,V1,V2):- in(V0,V3,V2),add(V3,V4,V1),
                in(V0,V5,V6),lt(V1,V5).
```

POPPER with our relational representation learns the following programs on the *ARC* domain:

```
% arc_a85d4709
out(V0,V1,V2,V3):- value_2(V3),in(V0,V6,V1,V4),
                   in(V0,V5,V2,V4),position_0(V5).
out(V0,V1,V2,V3):- value_3(V3),in(V0,V6,V1,V4),
                   in(V0,V5,V2,V4),rightmost_col(V0,V5).
out(V0,V1,V2,V3):- value_4(V3),in(V0,V5,V1,V4),
                   in(V0,V6,V2,V4),mid_row(V0,V6).

% arc_a699fb00
out(V0,V1,V2,V3):- value_2(V3),succ(V1,V6),
                   succ(V4,V1),in(V0,V4,V2,V5),
                   in(V0,V6,V2,V5).
out(V0,V1,V2,V3):- in(V0,V1,V2,V3).

% arc_94f9d214
out(V0,V1,V2,V3):- value_2(V3),empty(V0,V1,V2),
                   mid_row(V0,V4),add(V2,V4,V5),
                   empty(V0,V1,V5).

% arc_4258a5f9
out(V0,V1,V2,V3):- in(V0,V1,V2,V3).
out(V0,V1,V2,V3):- value_1(V3),succ(V2,V5),
                   succ(V1,V4),in(V0,V4,V5,V6).
out(V0,V1,V2,V3):- value_1(V3),succ(V4,V1),
                   in(V0,V4,V2,V5).
out(V0,V1,V2,V3):- value_1(V3),succ(V2,V5),
                   succ(V4,V1),in(V0,V4,V5,V6).
out(V0,V1,V2,V3):- value_1(V3),succ(V5,V1),
                   succ(V4,V2),in(V0,V5,V4,V6).
out(V0,V1,V2,V3):- value_1(V3),succ(V5,V2),
                   in(V0,V1,V5,V4).
out(V0,V1,V2,V3):- value_1(V3),succ(V2,V5),
                   in(V0,V1,V5,V4).
out(V0,V1,V2,V3):- value_1(V3),succ(V1,V5),
                   succ(V4,V2),in(V0,V5,V4,V6).

% arc_68b67ca3 (arc_67e8384a?)
out(V0,V1,V2,V3):- value_1(V3),succ(V1,V4),
                   in(V0,V4,V2,V5).

% arc_67e8384a
out(V0,V1,V2,V3):- position_5(V4),add(V2,V6,V4),
                   add(V1,V5,V4),in(V0,V5,V6,V3).
out(V0,V1,V2,V3):- in(V0,V1,V2,V3).
out(V0,V1,V2,V3):- position_5(V5),add(V2,V4,V5),
                   in(V0,V1,V4,V3).
out(V0,V1,V2,V3):- position_5(V5),add(V1,V4,V5),
                   in(V0,V4,V2,V3).

% arc_25ff71a9
out(V0,V1,V2,V3):- succ(V4,V2),in(V0,V1,V4,V3).

% arc_99b1bc43
out(V0,V1,V2,V3):- empty(V0,V1,V2),value_3(V3),
                   position_5(V6),add(V2,V6,V4),
                   in(V0,V1,V4,V5).
out(V0,V1,V2,V3):- value_3(V3),in(V0,V1,V2,V4),
                   position_5(V5),add(V2,V5,V6),
                   empty(V0,V1,V6).

% arc_d037b0a7
out(V0,V1,V2,V3):- in(V0,V1,V5,V3),add(V4,V5,V2),
                   empty(V0,V6,V2).

% arc_025d127b
out(V0,V1,V2,V3):- in(V0,V1,V2,V6),in(V0,V5,V2,V3),
                   in(V0,V5,V4,V3),lt(V4,V2),
                   empty(V0,V1,V4),empty(V0,V4,V5).
out(V0,V1,V2,V3):- succ(V4,V1),in(V0,V4,V2,V3),
                   in(V0,V1,V5,V3).

% arc_ed36ccf7
out(V0,V1,V2,V3):- rightmost_col(V0,V4),add(V2,V5,V4),
                   in(V0,V5,V1,V3).

% arc_b60334d2
out(V0,V1,V2,V3):- value_1(V3),succ(V2,V4),
                   in(V0,V1,V4,V5).
out(V0,V1,V2,V3):- succ(V1,V5),succ(V2,V4),
                   in(V0,V5,V4,V3).
out(V0,V1,V2,V3):- succ(V5,V1),succ(V4,V2),
                   in(V0,V5,V4,V3).
out(V0,V1,V2,V3):- value_1(V3),succ(V1,V4),
                   in(V0,V4,V2,V5).
out(V0,V1,V2,V3):- value_1(V3),succ(V4,V2),
                   in(V0,V1,V4,V5).
out(V0,V1,V2,V3):- succ(V1,V5),succ(V4,V2),
                   in(V0,V5,V4,V3).
out(V0,V1,V2,V3):- succ(V2,V4),succ(V5,V1),
                   in(V0,V5,V4,V3).
out(V0,V1,V2,V3):- value_1(V3),succ(V4,V1),
                   in(V0,V4,V2,V5).

% arc_74dd1130
out(V0,V1,V2,V3):- in(V0,V2,V1,V3).

% arc_ded97339
out(V0,V1,V2,V3):- in(V0,V1,V2,V3).
out(V0,V1,V2,V3):- in(V0,V1,V6,V3),lt(V2,V6),
                   in(V0,V1,V5,V3),add(V4,V5,V2).
out(V0,V1,V2,V3):- in(V0,V6,V2,V3),add(V4,V6,V1),
                   in(V0,V5,V2,V3),lt(V1,V5).
```

```
% arc_6150a2bd
out(V0,V1,V2,V3):- bottom_row(V0,V4),add(V2,V6,V4),
                   add(V1,V5,V4),in(V0,V5,V6,V3).

% arc_3bd67248
out(V0,V1,V2,V3):- in(V0,V1,V2,V3).
out(V0,V1,V2,V3):- value_4(V3),bottom_row(V0,V2),
                   empty(V0,V1,V4).
out(V0,V1,V2,V3):- value_2(V3),add(V1,V2,V5),
                   bottom_row(V0,V5),empty(V0,V1,V4).

% arc_e9afcf9a
out(V0,V1,V2,V3):- position_0(V2),in(V0,V1,V5,V3),
                   add(V5,V6,V4),add(V4,V6,V1).
out(V0,V1,V2,V3):- succ(V5,V2),in(V0,V4,V5,V3),
                   add(V2,V6,V4),add(V4,V6,V1),
                   succ(V6,V4).
out(V0,V1,V2,V3):- add(V1,V2,V4),in(V0,V4,V2,V3),
                   add(V2,V6,V5),add(V5,V6,V4).

% arc_2281f1f4
out(V0,V1,V2,V3):- value_2(V3),rightmost_col(V0,V5),
                   in(V0,V5,V2,V6),in(V0,V1,V4,V6),
                   empty(V0,V5,V4).
out(V0,V1,V2,V3):- in(V0,V1,V2,V3).

% arc_68b16354
out(V0,V1,V2,V3):- rightmost_col(V0,V5),add(V2,V4,V5),
                   in(V0,V1,V4,V3).

% arc_6430c8c4
out(V0,V1,V2,V3):- value_3(V3),empty(V0,V1,V2),
                   position_5(V4),add(V2,V4,V5),
                   empty(V0,V1,V5).

% arc_a5f85a15
out(V0,V1,V2,V3):- value_4(V3),in(V0,V1,V2,V6),
                   succ(V4,V5),add(V4,V5,V1).
out(V0,V1,V2,V3):- in(V0,V1,V2,V3),add(V1,V2,V6),
                   add(V4,V5,V6),add(V2,V4,V5).

% arc_539a4f51
out(V0,V1,V2,V3):- empty(V0,V1,V2),in(V0,V4,V5,V3),
                   add(V2,V5,V6),add(V4,V6,V2).
out(V0,V1,V2,V3):- position_6(V5),add(V4,V5,V2),
                   in(V0,V1,V6,V3).
out(V0,V1,V2,V3):- width(V0,V2),in(V0,V4,V5,V3),
                   add(V4,V6,V1),add(V1,V5,V6).
out(V0,V1,V2,V3):- position_5(V1),in(V0,V4,V5,V3),
                   add(V2,V5,V6),add(V4,V6,V2).
out(V0,V1,V2,V3):- lt(V2,V1),position_5(V6),
                   lt(V6,V1),in(V0,V4,V5,V3).
out(V0,V1,V2,V3):- in(V0,V1,V2,V3).
out(V0,V1,V2,V3):- position_5(V6),lt(V6,V2),
                   in(V0,V4,V5,V3),lt(V4,V1).

% arc_6d0aefbc
out(V0,V1,V2,V3):- in(V0,V1,V2,V3).
out(V0,V1,V2,V3):- position_5(V5),add(V1,V4,V5),
                   in(V0,V4,V2,V3).

% arc_623ea044
out(V0,V1,V2,V3):- empty(V0,V2,V1),in(V0,V4,V5,V3),
                   add(V2,V4,V6),add(V1,V5,V6).

out(V0,V1,V2,V3):- empty(V0,V1,V2),add(V1,V2,V4),
                   in(V0,V5,V6,V3),add(V5,V6,V4).
out(V0,V1,V2,V3):- in(V0,V1,V2,V3).

% arc_d13f3404
out(V0,V1,V2,V3):- in(V0,V5,V4,V3),add(V4,V6,V2),
                   add(V5,V6,V1).

% arc_29c11459
out(V0,V1,V2,V3):- value_5(V3),position_5(V1),
                   in(V0,V4,V2,V5).
out(V0,V1,V2,V3):- in(V0,V4,V2,V3),add(V4,V5,V1),
                   empty(V0,V1,V5).
out(V0,V1,V2,V3):- in(V0,V4,V2,V3),add(V1,V5,V4),
                   empty(V0,V2,V5).

% arc_6fa7a44f
out(V0,V1,V2,V3):- in(V0,V1,V2,V3).
out(V0,V1,V2,V3):- position_5(V5),add(V2,V4,V5),
                   in(V0,V1,V4,V3).

% arc_97999447
out(V0,V1,V2,V3):- value_5(V3),in(V0,V4,V2,V6),
                   add(V4,V5,V1),position_7(V5).
out(V0,V1,V2,V3):- value_5(V3),in(V0,V4,V2,V5),
                   succ(V4,V1).
out(V0,V1,V2,V3):- value_5(V3),in(V0,V4,V2,V6),
                   add(V4,V5,V1),position_9(V5).
out(V0,V1,V2,V3):- value_5(V3),position_5(V4),
                   add(V4,V5,V1),empty(V0,V1,V5),
                   in(V0,V5,V2,V6).
out(V0,V1,V2,V3):- in(V0,V6,V2,V3),empty(V0,V1,V5),
                   add(V5,V6,V4),add(V4,V5,V1).
out(V0,V1,V2,V3):- value_5(V3),in(V0,V4,V2,V6),
                   add(V4,V5,V1),position_3(V5).

% arc_f25ffba3
out(V0,V1,V2,V3):- in(V0,V1,V2,V3).
out(V0,V1,V2,V3):- position_9(V5),add(V2,V4,V5),
                   in(V0,V1,V4,V3).

% arc_c9f8e694
out(V0,V1,V2,V3):- in(V0,V1,V2,V5),position_0(V4),
                   in(V0,V4,V2,V3).

% arc_e3497940
out(V0,V1,V2,V3):- in(V0,V1,V2,V3),position_3(V4),
                   add(V1,V5,V4).
out(V0,V1,V2,V3):- rightmost_col(V0,V4),add(V1,V5,V4),
                   lt(V1,V5),in(V0,V5,V2,V3).

% arc_c3f564a4
out(V0,V1,V2,V3):- in(V0,V5,V1,V3),in(V0,V4,V5,V6),
                   in(V0,V2,V4,V6).

% arc_f76d97a5
out(V0,V1,V2,V3):- value_5(V5),different(V3,V5),
                   in(V0,V1,V2,V5),in(V0,V6,V4,V3).

% arc_3af2c5a8
out(V0,V1,V2,V3):- in(V0,V1,V2,V3).
out(V0,V1,V2,V3):- position_7(V5),add(V1,V4,V5),
                   lt(V4,V2),in(V0,V4,V6,V3).
out(V0,V1,V2,V3):- position_7(V5),add(V1,V4,V5),
                   in(V0,V4,V2,V3).
```

```
out(V0,V1,V2,V3):- position_5(V5),add(V2,V4,V5),
                   in(V0,V1,V4,V3).

% arc_dc433765
out(V0,V1,V2,V3):- in(V0,V1,V2,V3),value_4(V3).
out(V0,V1,V2,V3):- position_2(V1),succ(V5,V2),
                   position_1(V5),in(V0,V5,V4,V3).
out(V0,V1,V2,V3):- position_6(V2),in(V0,V1,V4,V3),
                   succ(V2,V4).
out(V0,V1,V2,V3):- position_2(V6),in(V0,V2,V6,V3),
                   in(V0,V1,V5,V4),lt(V6,V5).
out(V0,V1,V2,V3):- mid_col(V0,V1),position_1(V2),
                   value_3(V3),add(V1,V4,V2).

% arc_62c24649
out(V0,V1,V2,V3):- position_5(V5),add(V1,V4,V5),
                   in(V0,V4,V2,V3).
out(V0,V1,V2,V3):- position_5(V5),add(V2,V4,V5),
                   in(V0,V1,V4,V3).
out(V0,V1,V2,V3):- in(V0,V1,V2,V3).
out(V0,V1,V2,V3):- position_5(V6),add(V1,V4,V6),
                   add(V2,V5,V6),in(V0,V4,V5,V3).

% arc_fafffa47
out(V0,V1,V2,V3):- empty(V0,V1,V2),value_2(V3),
                   position_3(V4),add(V2,V4,V5),
                   empty(V0,V1,V5).

% arc_1b2d62fb
out(V0,V1,V2,V3):- value_8(V3),empty(V0,V1,V2),
                   bottom_row(V0,V5),add(V1,V5,V4),
                   empty(V0,V4,V2).

% arc_4347f46a
out(V0,V1,V2,V3):- in(V0,V1,V2,V3),succ(V2,V5),
                   succ(V4,V1),empty(V0,V4,V5).
out(V0,V1,V2,V3):- in(V0,V1,V2,V3),succ(V1,V4),
                   succ(V5,V2),empty(V0,V4,V5).

% arc_253bf280
out(V0,V1,V2,V3):- value_3(V3),in(V0,V5,V2,V4),
                   lt(V1,V5),in(V0,V6,V2,V4),lt(V6,V1).
out(V0,V1,V2,V3):- in(V0,V1,V2,V3).
out(V0,V1,V2,V3):- value_3(V3),in(V0,V1,V6,V4),
                   lt(V2,V6),in(V0,V1,V5,V4),lt(V5,V2).

% arc_ce4f8723
out(V0,V1,V2,V3):- value_3(V3),in(V0,V1,V2,V4),
                   in(V0,V5,V1,V4).
out(V0,V1,V2,V3):- value_3(V3),position_5(V4),
                   add(V2,V4,V5),in(V0,V1,V5,V6).

% arc_3c9b0459
out(V0,V1,V2,V3):- bottom_row(V0,V5),add(V2,V4,V5),
                   add(V1,V6,V5),in(V0,V6,V4,V3).

% arc_2dee498d
out(V0,V1,V2,V3):- in(V0,V1,V2,V3),in(V0,V4,V1,V5).

% arc_6d75e8bb
out(V0,V1,V2,V3):- empty(V0,V1,V2),value_2(V3),
                   in(V0,V5,V2,V4),in(V0,V1,V6,V4).
out(V0,V1,V2,V3):- in(V0,V1,V2,V3).

% arc_22eb0ac0
out(V0,V1,V2,V3):- in(V0,V6,V2,V3),lt(V1,V6),
                   in(V0,V4,V2,V3),add(V4,V5,V1).
out(V0,V1,V2,V3):- in(V0,V1,V2,V3).

% arc_25d8a9c8
out(V0,V1,V2,V3):- value_5(V3),rightmost_col(V0,V4),
                   in(V0,V1,V2,V5),in(V0,V4,V2,V5),
                   add(V2,V6,V4),lt(V6,V4),
                   in(V0,V6,V2,V5).
out(V0,V1,V2,V3):- value_5(V3),value_4(V5),
                   in(V0,V1,V4,V5),in(V0,V4,V2,V5).

% arc_c9e6f938
out(V0,V1,V2,V3):- in(V0,V1,V2,V3).
out(V0,V1,V2,V3):- position_5(V5),add(V1,V4,V5),
                   in(V0,V4,V2,V3).

% arc_27a28665
out(V0,V1,V2,V3):- value_1(V3),succ(V2,V4),
                   add(V1,V5,V2),add(V2,V4,V6),
                   empty(V0,V6,V4).
out(V0,V1,V2,V3):- value_3(V3),empty(V0,V2,V1),
                   succ(V1,V4),empty(V0,V2,V4).
out(V0,V1,V2,V3):- value_6(V3),succ(V1,V5),
                   add(V2,V5,V6),in(V0,V5,V6,V4),
                   in(V0,V1,V6,V4).
out(V0,V1,V2,V3):- value_2(V3),succ(V1,V5),
                   empty(V0,V5,V2),add(V1,V4,V6),
                   add(V2,V6,V4).

% arc_0520fde7
out(V0,V1,V2,V3):- value_2(V3),position_4(V5),
                   add(V1,V5,V4),in(V0,V1,V2,V6),
                   in(V0,V4,V2,V6).

% arc_a61f2674
out(V0,V1,V2,V3):- value_2(V3),in(V0,V1,V2,V4),
                   position_5(V5),empty(V0,V1,V5).
out(V0,V1,V2,V3):- value_1(V3),position_1(V6),
                   in(V0,V1,V6,V4),in(V0,V5,V2,V4).

% arc_4c4377d9
out(V0,V1,V2,V3):- position_3(V5),add(V4,V5,V2),
                   in(V0,V1,V4,V3).
out(V0,V1,V2,V3):- position_2(V5),add(V2,V4,V5),
                   in(V0,V1,V4,V3).

% arc_91413438
out(V0,V1,V2,V3):- position_2(V6),add(V5,V6,V1),
                   in(V0,V4,V2,V3),lt(V4,V5).
out(V0,V1,V2,V3):- position_4(V2),in(V0,V4,V1,V3).
out(V0,V1,V2,V3):- position_5(V6),add(V2,V5,V6),
                   empty(V0,V1,V5),in(V0,V4,V5,V3),
                   in(V0,V5,V1,V4).
out(V0,V1,V2,V3):- in(V0,V1,V2,V3).
out(V0,V1,V2,V3):- in(V0,V5,V4,V3),lt(V4,V1),
                   succ(V5,V4),add(V4,V6,V2),lt(V4,V6).

% arc_d10ecb37
out(V0,V1,V2,V3):- in(V0,V1,V2,V3),position_2(V4),
                   lt(V2,V4),lt(V1,V4).

% arc_eb281b96
out(V0,V1,V2,V3):- height(V0,V4),lt(V4,V2),
                   in(V0,V1,V5,V3).
```

```
out(V0,V1,V2,V3):- in(V0,V1,V2,V3).
out(V0,V1,V2,V3):- height(V0,V2),mid_row(V0,V4),
                   in(V0,V1,V4,V3).

% arc_c8f0f002
out(V0,V1,V2,V3):- in(V0,V1,V2,V3),value_7(V4),
                   different(V3,V4).
out(V0,V1,V2,V3):- value_5(V3),value_7(V4),
                   in(V0,V1,V2,V4).

% arc_ce22a75a
out(V0,V1,V2,V3):- value_1(V3),in(V0,V1,V2,V4).
out(V0,V1,V2,V3):- value_1(V3),succ(V5,V2),
                   succ(V1,V6),in(V0,V6,V5,V4).
out(V0,V1,V2,V3):- value_1(V3),succ(V2,V4),
                   in(V0,V1,V4,V5).
out(V0,V1,V2,V3):- value_1(V3),succ(V5,V1),
                   in(V0,V5,V2,V4).
out(V0,V1,V2,V3):- value_1(V3),succ(V6,V1),
                   succ(V5,V2),in(V0,V6,V5,V4).
out(V0,V1,V2,V3):- value_1(V3),succ(V2,V5),
                   succ(V1,V6),in(V0,V6,V5,V4).
out(V0,V1,V2,V3):- value_1(V3),succ(V4,V2),
                   in(V0,V1,V4,V5).
out(V0,V1,V2,V3):- value_1(V3),succ(V1,V5),
                   in(V0,V5,V2,V4).
out(V0,V1,V2,V3):- value_1(V3),succ(V6,V1),
                   succ(V2,V5),in(V0,V6,V5,V4).

% arc_e98196ab
out(V0,V1,V2,V3):- position_6(V5),add(V2,V5,V4),
                   in(V0,V1,V4,V3).
out(V0,V1,V2,V3):- in(V0,V1,V2,V3),mid_col(V0,V4),
                   lt(V2,V4).

% arc_7fe24cdd
out(V0,V1,V2,V3):- position_5(V4),add(V2,V5,V4),
                   in(V0,V5,V1,V3).
out(V0,V1,V2,V3):- in(V0,V1,V2,V3).
out(V0,V1,V2,V3):- position_5(V5),add(V1,V4,V5),
                   in(V0,V2,V4,V3).
out(V0,V1,V2,V3):- position_5(V6),add(V2,V5,V6),
                   add(V1,V4,V6),in(V0,V4,V5,V3).

% arc_d23f8c26
out(V0,V1,V2,V3):- in(V0,V1,V2,V3),mid_col(V0,V1).

% arc_82819916
out(V0,V1,V2,V3):- in(V0,V1,V4,V5),in(V0,V6,V4,V5),
                   in(V0,V6,V2,V3).

% arc_42a50994
out(V0,V1,V2,V3):- in(V0,V1,V2,V3),succ(V2,V4),
                   succ(V6,V1),in(V0,V6,V4,V5).
out(V0,V1,V2,V3):- in(V0,V1,V2,V3),succ(V2,V4),
                   succ(V1,V6),in(V0,V6,V4,V5).
out(V0,V1,V2,V3):- in(V0,V1,V2,V3),succ(V1,V4),
                   succ(V5,V2),in(V0,V4,V5,V6).
out(V0,V1,V2,V3):- succ(V2,V5),in(V0,V1,V5,V3),
                   in(V0,V1,V2,V4).
out(V0,V1,V2,V3):- in(V0,V1,V2,V3),succ(V4,V1),
                   in(V0,V4,V2,V5).
out(V0,V1,V2,V3):- in(V0,V1,V2,V3),succ(V1,V4),
                   in(V0,V4,V2,V5).
out(V0,V1,V2,V3):- succ(V5,V2),in(V0,V1,V5,V3),
                   in(V0,V1,V2,V4).
out(V0,V1,V2,V3):- in(V0,V1,V2,V3),succ(V4,V1),
                   succ(V5,V2),in(V0,V4,V5,V6).

% arc_67a3c6ac
out(V0,V1,V2,V3):- rightmost_col(V0,V5),
                   add(V1,V4,V5),in(V0,V4,V2,V3).

% arc_feca6190
out(V0,V1,V2,V3):- succ(V5,V2),in(V0,V4,V6,V3),
                   lt(V6,V1),lt(V6,V5).
out(V0,V1,V2,V3):- position_4(V1),empty(V0,V1,V5),
                   empty(V0,V5,V4),in(V0,V2,V4,V3).
out(V0,V1,V2,V3):- in(V0,V4,V5,V3),succ(V1,V6),lt(V6,V2).
out(V0,V1,V2,V3):- position_7(V6),lt(V2,V6),
                   lt(V6,V1),in(V0,V4,V5,V3).

% arc_a416b8f3
out(V0,V1,V2,V3):- width(V0,V5),add(V4,V5,V1),
                   in(V0,V4,V2,V3).
out(V0,V1,V2,V3):- in(V0,V1,V2,V3).

% arc_b1948b0a
out(V0,V1,V2,V3):- value_7(V3),in(V0,V1,V2,V3).
out(V0,V1,V2,V3):- value_2(V3),in(V0,V1,V2,V4),
                   value_6(V4).

% arc_963e52fc
out(V0,V1,V2,V3):- position_9(V5),add(V4,V5,V1),
                   in(V0,V4,V2,V3).
out(V0,V1,V2,V3):- position_6(V5),add(V4,V5,V1),
                   in(V0,V4,V2,V3).
out(V0,V1,V2,V3):- in(V0,V1,V2,V3).

% arc_3618c87e
out(V0,V1,V2,V3):- position_3(V5),add(V5,V6,V2),
                   add(V4,V6,V5),in(V0,V1,V4,V3).
out(V0,V1,V2,V3):- in(V0,V1,V2,V3),empty(V0,V1,V5),
                   in(V0,V4,V5,V3).

% arc_b8825c91
out(V0,V1,V2,V3):- in(V0,V1,V2,V3),value_4(V4),
                   different(V3,V4).
out(V0,V1,V2,V3):- bottom_row(V0,V4),add(V2,V5,V4),
                   in(V0,V5,V1,V3),in(V0,V4,V6,V3).
out(V0,V1,V2,V3):- value_5(V3),bottom_row(V0,V4),
                   add(V1,V5,V4),in(V0,V5,V2,V3).
out(V0,V1,V2,V3):- value_3(V3),bottom_row(V0,V4),
                   add(V2,V5,V4),in(V0,V5,V1,V3).
out(V0,V1,V2,V3):- in(V0,V2,V1,V3),value_4(V4),
                   different(V3,V4).
out(V0,V1,V2,V3):- value_8(V3),bottom_row(V0,V5),
                   add(V1,V4,V5),in(V0,V4,V2,V3).

% arc_6f8cd79b
out(V0,V1,V2,V3):- value_8(V3),bottom_row(V0,V2),
                   empty(V0,V1,V4).
out(V0,V1,V2,V3):- value_8(V3),position_0(V2),
                   empty(V0,V1,V4).
out(V0,V1,V2,V3):- position_0(V1),value_8(V3),
                   empty(V0,V4,V2).
out(V0,V1,V2,V3):- rightmost_col(V0,V1),value_8(V3),
                   empty(V0,V4,V2).

% arc_0dfd9992
```

```
out(V0,V1,V2,V3):- in(V0,V6,V1,V3),in(V0,V5,V2,V4),
                   in(V0,V5,V6,V4).

% arc_9dfd6313
out(V0,V1,V2,V3):- in(V0,V2,V1,V3).

% arc_d511f180
out(V0,V1,V2,V3):- in(V0,V1,V2,V3),value_4(V3).
out(V0,V1,V2,V3):- in(V0,V1,V2,V3),value_2(V3).
out(V0,V1,V2,V3):- value_8(V3),value_5(V4),
                   in(V0,V1,V2,V4).
out(V0,V1,V2,V3):- in(V0,V1,V2,V3),value_1(V3).
out(V0,V1,V2,V3):- in(V0,V1,V2,V3),value_9(V3).
out(V0,V1,V2,V3):- in(V0,V1,V2,V3),value_7(V3).
out(V0,V1,V2,V3):- in(V0,V1,V2,V3),value_6(V3).
out(V0,V1,V2,V3):- in(V0,V1,V2,V3),value_3(V3).
out(V0,V1,V2,V3):- value_5(V3),value_8(V4),
                   in(V0,V1,V2,V4).

% arc_0d3d703e
out(V0,V1,V2,V3):- value_5(V3),in(V0,V1,V2,V4),
                   value_1(V4).
out(V0,V1,V2,V3):- value_1(V3),value_5(V4),
                   in(V0,V1,V2,V4).
out(V0,V1,V2,V3):- value_6(V3),value_2(V4),
                   in(V0,V1,V2,V4).
out(V0,V1,V2,V3):- value_8(V3),value_9(V4),
                   in(V0,V1,V2,V4).
out(V0,V1,V2,V3):- value_9(V3),in(V0,V1,V2,V4),
                   value_8(V4).
out(V0,V1,V2,V3):- value_3(V3),in(V0,V1,V2,V4),
                   value_4(V4).
out(V0,V1,V2,V3):- value_4(V3),in(V0,V1,V2,V4),
                   value_3(V4).
out(V0,V1,V2,V3):- value_2(V3),in(V0,V1,V2,V4),
                   value_6(V4).

% arc_5bd6f4ac
out(V0,V1,V2,V3):- position_3(V4),lt(V2,V4),
                   add(V1,V4,V5),add(V4,V5,V6),
                   in(V0,V6,V2,V3).

% arc_a79310a0
out(V0,V1,V2,V3):- value_2(V3),succ(V4,V2),
                   in(V0,V1,V4,V5).

% arc_496994bd
out(V0,V1,V2,V3):- bottom_row(V0,V4),add(V2,V5,V4),
                   in(V0,V1,V5,V3).
out(V0,V1,V2,V3):- in(V0,V1,V2,V3).

% arc_f2829549
out(V0,V1,V2,V3):- empty(V0,V1,V2),value_3(V3),
                   position_4(V5),add(V1,V5,V4),
                   empty(V0,V4,V2).

% arc_aabf363d
out(V0,V1,V2,V3):- position_0(V5),empty(V0,V5,V2),
                   in(V0,V1,V2,V4),in(V0,V5,V6,V3).

% arc_bbc9ae5d
out(V0,V1,V2,V3):- add(V1,V5,V2),in(V0,V4,V6,V3).
out(V0,V1,V2,V3):- add(V2,V5,V1),in(V0,V5,V4,V3).

% arc_22168020
```

```
out(V0,V1,V2,V3):- in(V0,V4,V2,V3),in(V0,V1,V5,V3),
                   add(V2,V6,V5).

% arc_7b7f7511
out(V0,V1,V2,V3):- in(V0,V1,V2,V3),in(V0,V2,V1,V4).
out(V0,V1,V2,V3):- in(V0,V1,V2,V3),width(V0,V2).

% arc_8be77c9e
out(V0,V1,V2,V3):- in(V0,V1,V2,V3).
out(V0,V1,V2,V3):- position_5(V4),add(V2,V5,V4),
                   in(V0,V1,V5,V3).

% arc_dae9d2b5
out(V0,V1,V2,V3):- value_6(V3),in(V0,V1,V2,V5),
                   in(V0,V6,V1,V4).
out(V0,V1,V2,V3):- value_6(V3),height(V0,V5),
                   add(V1,V5,V6),in(V0,V6,V2,V4).
```

POPPER with our relational representation learns the following programs on the *listfunctions* domain:

```
% list_001
out(V0,V1,V2):- c0(V1),c2(V3),in(V0,V3,V2).

% list_002
out(V0,V1,V2):- c0(V1),c2(V3),in(V0,V3,V2).

% list_003
out(V0,V1,V2):- c0(V1),c6(V3),in(V0,V3,V2).

% list_004
out(V0,V1,V2):- c0(V1),c6(V3),in(V0,V3,V2).

% list_005
out(V0,V1,V2):- c0(V1),in(V0,V1,V3),in(V0,V3,V2).

% list_006
out(V0,V1,V2):- in(V0,V1,V2),c2(V3),lt(V1,V3).

% list_007
out(V0,V1,V2):- in(V0,V1,V2),c2(V3),lt(V1,V3).

% list_008
out(V0,V1,V2):- in(V0,V1,V2),c5(V3),add(V1,V4,V3).

% list_009
out(V0,V1,V2):- in(V0,V1,V2),c5(V3),add(V1,V4,V3).

% list_010
out(V0,V1,V2):- c0(V4),in(V0,V4,V5),lt(V1,V5),
                succ(V1,V3),in(V0,V3,V2).

% list_011
out(V0,V1,V2):- c2(V5),succ(V1,V4),
                in(V0,V4,V2),add(V1,V3,V5).

% list_012
out(V0,V1,V2):- c2(V4),add(V1,V3,V4),
                succ(V1,V5),in(V0,V5,V2).

% list_013
out(V0,V1,V2):- c2(V4),add(V1,V4,V3),
                in(V0,V3,V2),c7(V5),lt(V3,V5).

% list_014
out(V0,V1,V2):- c2(V5),add(V1,V5,V3),
```

```
                    in(V0,V3,V2),c5(V4),lt(V1,V4).

% list_016
out(V0,V1,V2):- c1(V1),c8(V2).
out(V0,V1,V2):- in(V0,V1,V2),c2(V4),add(V3,V4,V1).
out(V0,V1,V2):- in(V0,V1,V2),c0(V1).

% list_017
out(V0,V1,V2):- in(V0,V1,V2),c2(V4),add(V3,V4,V1).
out(V0,V1,V2):- c0(V1),in(V0,V1,V2).
out(V0,V1,V2):- c8(V2),c1(V1),in(V0,V1,V3).

% list_018
out(V0,V1,V2):- in(V0,V1,V2),c6(V3),add(V3,V4,V1).
out(V0,V1,V2):- c5(V1),c3(V2).
out(V0,V1,V2):- in(V0,V1,V2),c4(V4),add(V1,V3,V4).

% list_019
out(V0,V1,V2):- in(V0,V1,V2),c6(V4),add(V3,V4,V1).
out(V0,V1,V2):- in(V0,V1,V2),c4(V3),add(V1,V4,V3).
out(V0,V1,V2):- c3(V2),c5(V1),in(V0,V1,V3).

% list_020
out(V0,V1,V2):- end_pos(V0,V5),succ(V3,V5),
                add(V1,V3,V4),in(V0,V4,V2).
out(V0,V1,V2):- in(V0,V1,V2),succ(V3,V1).

% list_021
out(V0,V1,V2):- succ(V4,V1),in(V0,V4,V2),lt(V3,V4).
out(V0,V1,V2):- c8(V2),c1(V1).
out(V0,V1,V2):- c0(V1),in(V0,V1,V2).

% list_022
out(V0,V1,V2):- succ(V4,V1),in(V0,V4,V2),lt(V3,V4).
out(V0,V1,V2):- c5(V2),c1(V1).
out(V0,V1,V2):- c0(V1),in(V0,V1,V2).

% list_023
out(V0,V1,V2):- c1(V1),c8(V2),end_pos(V0,V3),
                add(V3,V5,V2),add(V3,V4,V5).
out(V0,V1,V2):- c5(V2),c1(V1),c4(V3),in(V0,V3,V4).
out(V0,V1,V2):- succ(V3,V1),in(V0,V3,V2),lt(V4,V3).
out(V0,V1,V2):- c0(V1),in(V0,V1,V2).

% list_024
out(V0,V1,V2):- c5(V2),c1(V1),c0(V4),
                in(V0,V4,V3),add(V2,V5,V3).
out(V0,V1,V2):- c1(V1),c8(V2),in(V0,V3,V5),
                add(V4,V5,V2),add(V3,V5,V4).
out(V0,V1,V2):- in(V0,V1,V2),c0(V1).
out(V0,V1,V2):- succ(V3,V1),in(V0,V3,V2),lt(V4,V3).

% list_025
out(V0,V1,V2):- succ(V1,V4),in(V0,V4,V2),
                succ(V3,V1).
out(V0,V1,V2):- in(V0,V1,V2),c0(V1).

% list_026
out(V0,V1,V2):- in(V0,V1,V2),c2(V3),lt(V1,V3).
out(V0,V1,V2):- succ(V1,V5),in(V0,V5,V2),
                succ(V4,V1),lt(V3,V4).

% list_028
out(V0,V1,V2):- c3(V3),succ(V1,V4),
                in(V0,V4,V2),add(V3,V5,V4).

out(V0,V1,V2):- in(V0,V1,V2),c2(V5),lt(V1,V5),
                in(V0,V5,V3),add(V3,V4,V5).
out(V0,V1,V2):- c0(V1),in(V0,V1,V2).
out(V0,V1,V2):- succ(V1,V5),in(V0,V5,V2),
                add(V4,V5,V2),succ(V3,V1).

% list_029
out(V0,V1,V2):- c2(V3),add(V1,V3,V4),in(V0,V4,V2).

% list_030
out(V0,V1,V2):- in(V0,V1,V2),c2(V5),
                add(V1,V5,V4),in(V0,V4,V3).

% list_033
out(V0,V1,V2):- c3(V3),in(V0,V5,V2),
                add(V4,V5,V3),add(V1,V5,V4).
out(V0,V1,V2):- c0(V1),c3(V3),in(V0,V3,V2).
out(V0,V1,V2):- in(V0,V1,V2),add(V4,V5,V1),
                add(V3,V5,V4),lt(V3,V5).

% list_034
out(V0,V1,V2):- c2(V4),add(V1,V5,V4),
                lt(V5,V4),succ(V5,V3),in(V0,V3,V2).
out(V0,V1,V2):- in(V0,V1,V2),c0(V1).
out(V0,V1,V2):- in(V0,V1,V2),c3(V3),add(V3,V4,V1).

% list_035
out(V0,V1,V2):- in(V0,V1,V2),c4(V3),add(V3,V4,V1).
out(V0,V1,V2):- c3(V4),add(V1,V3,V4),in(V0,V3,V2).

% list_037
out(V0,V1,V2):- end_pos(V0,V1),c3(V2).
out(V0,V1,V2):- in(V0,V1,V2).

% list_038
out(V0,V1,V2):- in(V0,V1,V2).
out(V0,V1,V2):- c9(V2),end_pos(V0,V1).

% list_039
out(V0,V1,V2):- c9(V1),end_pos(V0,V2),add(V1,V3,V2).
out(V0,V1,V2):- c3(V1),end_pos(V0,V2),add(V2,V3,V1).
out(V0,V1,V2):- in(V0,V1,V2).

% list_040
out(V0,V1,V2):- end_pos(V0,V1),c9(V2),in(V0,V3,V2).
out(V0,V1,V2):- in(V0,V1,V2).
out(V0,V1,V2):- c3(V2),end_pos(V0,V1),in(V0,V3,V2).

% list_041
out(V0,V1,V2):- c0(V1),c9(V2).

% list_042
out(V0,V1,V2):- c2(V2),c1(V1).
out(V0,V1,V2):- c0(V1),c5(V2).

% list_043
out(V0,V1,V2):- c3(V2),c4(V1).
out(V0,V1,V2):- c0(V2),c3(V1).
out(V0,V1,V2):- c7(V2),c2(V1).
out(V0,V1,V2):- c8(V2),c0(V1).
out(V0,V1,V2):- c2(V2),c1(V1).

% list_044
out(V0,V1,V2):- lt(V1,V2),c1(V2).
out(V0,V1,V2):- c7(V1),c0(V2).
```

```
out(V0,V1,V2):- c4(V1),c2(V2).
out(V0,V1,V2):- c9(V2),add(V2,V3,V1),add(V1,V4,V2).
out(V0,V1,V2):- c5(V2),add(V1,V3,V2),add(V2,V4,V1).
out(V0,V1,V2):- c9(V2),c1(V1).
out(V0,V1,V2):- c4(V2),c8(V1).
out(V0,V1,V2):- c2(V1),c4(V2).
out(V0,V1,V2):- c3(V2),add(V1,V3,V2),add(V2,V4,V1).
out(V0,V1,V2):- c8(V2),c6(V1).

% list_045
out(V0,V1,V2):- in(V0,V1,V2).

% list_046
out(V0,V1,V2):- c0(V1),c7(V2).
out(V0,V1,V2):- succ(V3,V1),in(V0,V3,V2).

% list_047
out(V0,V1,V2):- c3(V2),c2(V1).
out(V0,V1,V2):- c6(V2),c1(V1).
out(V0,V1,V2):- c0(V1),c9(V2).
out(V0,V1,V2):- c3(V1),c8(V2).
out(V0,V1,V2):- c4(V1),c5(V2).
out(V0,V1,V2):- c5(V3),add(V3,V4,V1),in(V0,V4,V2).

% list_048
out(V0,V1,V2):- c0(V1),in(V0,V1,V2).

% list_049
out(V0,V1,V2):- succ(V1,V3),in(V0,V3,V2).

% list_050
out(V0,V1,V2):- succ(V3,V1),in(V0,V3,V2).
out(V0,V1,V2):- c0(V1),in(V0,V1,V2).

% list_051
out(V0,V1,V2):- c5(V5),add(V1,V4,V5),c0(V3),in(V0,V3,V2).
out(V0,V1,V2):- c5(V3),add(V3,V4,V1),in(V0,V4,V2).

% list_052
out(V0,V1,V2):- c0(V4),in(V0,V4,V2),c9(V5),add(V1,V3,V5).

% list_053
out(V0,V1,V2):- in(V0,V1,V2),c2(V4),add(V3,V4,V1).
out(V0,V1,V2):- c2(V4),lt(V1,V4),c0(V3),in(V0,V3,V2).

% list_054
out(V0,V1,V2):- c2(V4),in(V0,V4,V2),add(V1,V3,V4).
out(V0,V1,V2):- in(V0,V1,V2),c3(V4),add(V3,V4,V1).

% list_055
out(V0,V1,V2):- in(V0,V1,V2),c4(V3),add(V3,V4,V1).
out(V0,V1,V2):- c2(V4),add(V3,V4,V1),
                in(V0,V3,V2),lt(V3,V4).
out(V0,V1,V2):- c2(V4),lt(V1,V4),add(V1,V4,V3),
                in(V0,V3,V2).

% list_056
out(V0,V1,V2):- in(V0,V1,V2),c3(V4),add(V1,V3,V4).
out(V0,V1,V2):- succ(V1,V4),in(V0,V4,V2),
                c4(V3),add(V3,V5,V1).

% list_057
out(V0,V1,V2):- in(V0,V1,V2),c6(V3),lt(V1,V3).
out(V0,V1,V2):- c7(V4),add(V4,V5,V1),
                succ(V3,V1),in(V0,V3,V2).

out(V0,V1,V2):- c6(V1),c4(V2).

% list_058
out(V0,V1,V2):- c7(V4),add(V1,V4,V3),in(V0,V3,V2).

% list_059
out(V0,V1,V2):- c7(V1),c3(V3),in(V0,V3,V2).
out(V0,V1,V2):- in(V0,V1,V2),c2(V3),add(V1,V4,V3).
out(V0,V1,V2):- in(V0,V1,V2),c8(V3),add(V3,V4,V1).
out(V0,V1,V2):- c3(V1),c7(V3),in(V0,V3,V2).
out(V0,V1,V2):- in(V0,V1,V2),c6(V3),add(V1,V4,V3),
                lt(V4,V1).

% list_060
out(V0,V1,V2):- c2(V3),add(V1,V4,V3),in(V0,V4,V2).
out(V0,V1,V2):- c5(V1),in(V0,V3,V2),c6(V3).
out(V0,V1,V2):- c4(V1),in(V0,V1,V2).
out(V0,V1,V2):- c3(V1),c4(V2).

% list_061
out(V0,V1,V2):- end_pos(V0,V5),succ(V4,V5),
                add(V1,V4,V3),in(V0,V3,V2).

% list_062
out(V0,V1,V2):- in(V0,V1,V2),succ(V1,V4),in(V0,V4,V3).

% list_064
out(V0,V1,V2):- succ(V1,V5),in(V0,V5,V2),
                succ(V5,V3),in(V0,V3,V4).

% list_065
out(V0,V1,V2):- c9(V2),c0(V1).
out(V0,V1,V2):- succ(V3,V1),in(V0,V3,V2).
out(V0,V1,V2):- c7(V2),succ(V3,V1),end_pos(V0,V3).

% list_066
out(V0,V1,V2):- succ(V1,V4),end_pos(V0,V4),
                c0(V3),in(V0,V3,V2).
out(V0,V1,V2):- succ(V1,V3),in(V0,V3,V2).

% list_067
out(V0,V1,V2):- end_pos(V0,V4),succ(V1,V4),
                c0(V3),in(V0,V3,V2).
out(V0,V1,V2):- in(V0,V1,V2),succ(V4,V1),
                in(V0,V5,V3),lt(V1,V5).
out(V0,V1,V2):- end_pos(V0,V5),add(V1,V5,V4),
                succ(V3,V4),in(V0,V3,V2).

% list_068
out(V0,V1,V2):- c7(V2),end_pos(V0,V1).
out(V0,V1,V2):- c4(V2),c3(V3),add(V3,V4,V1),
                end_pos(V0,V4).
out(V0,V1,V2):- c3(V2),end_pos(V0,V3),succ(V3,V1).
out(V0,V1,V2):- c8(V2),c2(V3),add(V3,V4,V1),
                end_pos(V0,V4).
out(V0,V1,V2):- in(V0,V1,V2).
out(V0,V1,V2):- c3(V2),c4(V3),add(V3,V4,V1),
                end_pos(V0,V4).

% list_069
out(V0,V1,V2):- c7(V2),end_pos(V0,V3),
                add(V3,V4,V1),c4(V4).
out(V0,V1,V2):- c4(V2),c2(V1).
out(V0,V1,V2):- c0(V2),c3(V1).
out(V0,V1,V2):- c1(V2),c7(V4),add(V3,V4,V1),
```

```
                    end_pos(V0,V3).
out(V0,V1,V2):- c0(V1),c9(V2).
out(V0,V1,V2):- c2(V2),c5(V4),add(V3,V4,V1),
                    end_pos(V0,V3).
out(V0,V1,V2):- c3(V2),c1(V1).
out(V0,V1,V2):- c9(V2),end_pos(V0,V3),add(V3,V4,V1),
                    c6(V4).
out(V0,V1,V2):- c4(V3),add(V3,V4,V1),in(V0,V4,V2).

% list_070
out(V0,V1,V2):- end_pos(V0,V4),add(V3,V4,V1),in(V0,V3,V2).
out(V0,V1,V2):- in(V0,V1,V2).

% list_071
out(V0,V1,V2):- c2(V4),add(V3,V4,V2),in(V0,V1,V3).

% list_072
out(V0,V1,V2):- in(V0,V3,V2),succ(V3,V4),add(V3,V4,V1).
out(V0,V1,V2):- in(V0,V3,V2),add(V3,V4,V1),add(V4,V5,V3),
                    add(V3,V5,V4).

% list_073
out(V0,V1,V2):- in(V0,V1,V3),succ(V4,V2),add(V1,V3,V4).

% list_075
out(V0,V1,V2):- in(V0,V5,V2),add(V1,V4,V5),
                    add(V1,V3,V4),add(V3,V4,V1).

% list_077
out(V0,V1,V2):- c0(V1),end_pos(V0,V2).

% list_078
out(V0,V1,V2):- c0(V1),in(V0,V3,V2),succ(V3,V5),
                    in(V0,V5,V3),add(V3,V4,V2).
out(V0,V1,V2):- c0(V1),succ(V2,V4),end_pos(V0,V4),
                    in(V0,V3,V2),succ(V5,V3).
out(V0,V1,V2):- in(V0,V2,V4),add(V3,V4,V2),
                    add(V1,V4,V3),add(V3,V5,V4).

% list_080
out(V0,V1,V2):- succ(V1,V5),end_pos(V0,V4),
                    add(V3,V5,V4),in(V0,V3,V2).

% list_081
out(V0,V1,V2):- c0(V1),c2(V3),in(V0,V3,V2).

% list_082
out(V0,V1,V2):- c0(V1),c2(V3),in(V0,V3,V2).

% list_083
out(V0,V1,V2):- c0(V1),c6(V3),in(V0,V3,V2).

% list_084
out(V0,V1,V2):- c0(V1),c6(V3),in(V0,V3,V2).

% list_085
out(V0,V1,V2):- c0(V1),in(V0,V1,V3),in(V0,V3,V2).

% list_086
out(V0,V1,V2):- c3(V4),add(V1,V5,V4),add(V1,V3,V5),
                    in(V0,V3,V2).
out(V0,V1,V2):- c3(V1),c0(V3),in(V0,V3,V2).
out(V0,V1,V2):- in(V0,V1,V2),add(V3,V5,V1),
                    add(V5,V4,V3),lt(V4,V5).

% list_087
out(V0,V1,V2):- in(V0,V1,V2),c0(V1).
out(V0,V1,V2):- c3(V4),lt(V1,V4),succ(V3,V1),
                    add(V5,V1,V4),in(V0,V5,V2).
out(V0,V1,V2):- in(V0,V1,V2),c3(V4),add(V4,V3,V1).

% list_090
out(V0,V1,V2):- c1(V1),c42(V2).
out(V0,V1,V2):- c2(V1),c77(V2).
out(V0,V1,V2):- c36(V2),c4(V1).
out(V0,V1,V2):- c18(V2),c0(V1).
out(V0,V1,V2):- c20(V2),c3(V1).

% list_091
out(V0,V1,V2):- c3(V1),c23(V2).
out(V0,V1,V2):- c22(V2),c4(V1).
out(V0,V1,V2):- c1(V1),c99(V2).
out(V0,V1,V2):- c6(V1),c68(V2).
out(V0,V1,V2):- c81(V2),c0(V1).
out(V0,V1,V2):- c24(V2),c8(V1).
out(V0,V1,V2):- c30(V2),c7(V1).
out(V0,V1,V2):- c69(V2),c9(V1).
out(V0,V1,V2):- c5(V1),c75(V2).
out(V0,V1,V2):- c41(V2),c2(V1).

% list_092
out(V0,V1,V2):- c92(V2),c0(V1).
out(V0,V1,V2):- c5(V4),add(V3,V4,V1),in(V0,V3,V2).
out(V0,V1,V2):- c63(V2),c1(V1).
out(V0,V1,V2):- c55(V2),c4(V1).
out(V0,V1,V2):- c34(V2),c2(V1).
out(V0,V1,V2):- c18(V2),c3(V1).

% list_093
out(V0,V1,V2):- c0(V5),in(V0,V5,V2),c91(V3),add(V3,V1,V4).

% list_094
out(V0,V1,V2):- c2(V4),add(V3,V4,V1),lt(V3,V4),
                    in(V0,V3,V2).
out(V0,V1,V2):- c2(V4),lt(V1,V4),add(V1,V4,V3),
                    in(V0,V3,V2).
out(V0,V1,V2):- in(V0,V1,V2),c4(V3),add(V3,V4,V1).

% list_095
out(V0,V1,V2):- succ(V1,V5),in(V0,V5,V2),succ(V5,V3),
                    in(V0,V3,V4).

% list_096
out(V0,V1,V2):- succ(V3,V1),in(V0,V3,V2).
out(V0,V1,V2):- c98(V2),c0(V1).
out(V0,V1,V2):- c37(V2),end_pos(V0,V3),succ(V3,V1).

% list_097
out(V0,V1,V2):- c0(V1),c11(V2).
out(V0,V1,V2):- c7(V2),end_pos(V0,V3),add(V4,V3,V1),c4(V4).
out(V0,V1,V2):- c21(V2),c1(V1).
out(V0,V1,V2):- c89(V2),end_pos(V0,V3),add(V4,V3,V1),c5(V4).
out(V0,V1,V2):- c19(V2),c3(V1).
out(V0,V1,V2):- c0(V2),c6(V4),add(V4,V3,V1),end_pos(V0,V3).
out(V0,V1,V2):- c4(V3),add(V4,V3,V1),in(V0,V4,V2).
out(V0,V1,V2):- c57(V2),c7(V4),add(V4,V3,V1),end_pos(V0,V3).
out(V0,V1,V2):- c43(V2),c2(V1).

% list_098
out(V0,V1,V2):- succ(V4,V2),in(V0,V1,V3),add(V1,V3,V4).
```

```
% list_100
out(V0,V1,V2):- end_pos(V0,V5),succ(V1,V3),
                add(V3,V4,V5),in(V0,V4,V2).

% list_101
out(V0,V1,V2):- c9(V2),add(V3,V1,V4),add(V2,V3,V4).
out(V0,V1,V2):- c0(V1),c11(V2).
out(V0,V1,V2):- c24(V2),c2(V1).
out(V0,V1,V2):- c19(V2),c1(V1).
out(V0,V1,V2):- c64(V2),c8(V1).
out(V0,V1,V2):- c6(V1),c82(V2).
out(V0,V1,V2):- c7(V1),c0(V2).
out(V0,V1,V2):- c33(V2),c3(V1).
out(V0,V1,V2):- c5(V2),add(V3,V1,V4),add(V2,V3,V4).
out(V0,V1,V2):- c42(V2),c4(V1).

% list_102
out(V0,V1,V2):- in(V0,V1,V2).

% list_103
out(V0,V1,V2):- c0(V1),end_pos(V0,V2).

% list_114
out(V0,V1,V2):- in(V0,V3,V2),succ(V3,V5),
                add(V1,V4,V5),end_pos(V0,V4).
out(V0,V1,V2):- succ(V3,V1),in(V0,V3,V2).

% list_116
out(V0,V1,V2):- end_pos(V0,V3),in(V0,V5,V2),
                add(V1,V4,V3),succ(V5,V4).
out(V0,V1,V2):- end_pos(V0,V3),add(V4,V3,V1),
                in(V0,V5,V2),succ(V4,V5).

% list_120
out(V0,V1,V2):- in(V0,V1,V2),c0(V1).

% list_121
out(V0,V1,V2):- in(V0,V3,V2),add(V1,V5,V3),
                end_pos(V0,V4),succ(V5,V4).

% list_122
out(V0,V1,V2):- c0(V1),c2(V5),end_pos(V0,V4),
                add(V3,V5,V4),in(V0,V3,V2).

% list_126
out(V0,V1,V2):- succ(V1,V3),in(V0,V3,V2).

% list_127
out(V0,V1,V2):- in(V0,V1,V2),in(V0,V3,V4),lt(V1,V3).

% list_130
out(V0,V1,V2):- succ(V1,V4),in(V0,V4,V2),
                c0(V3),in(V0,V3,V5),lt(V1,V5).

% list_132
out(V0,V1,V2):- in(V0,V1,V2),c99(V3),add(V3,V1,V4).
out(V0,V1,V2):- succ(V1,V5),in(V0,V5,V2),lt(V4,V1),
                lt(V3,V4).

% list_133
out(V0,V1,V2):- c0(V1),in(V0,V1,V2).
out(V0,V1,V2):- in(V0,V4,V2),c4(V5),add(V5,V1,V4),
                succ(V3,V1).

% list_140
out(V0,V1,V2):- c0(V1),in(V0,V1,V2).
out(V0,V1,V2):- c1(V1),c9(V2).
out(V0,V1,V2):- in(V0,V1,V2),c2(V3),add(V4,V3,V1).

% list_145
out(V0,V1,V2):- in(V0,V1,V3),in(V0,V4,V2),c0(V4).

% list_146
out(V0,V1,V2):- in(V0,V1,V4),add(V2,V5,V4),succ(V1,V3),
                in(V0,V3,V5).
out(V0,V1,V2):- in(V0,V1,V4),add(V2,V4,V5),succ(V1,V3),
                in(V0,V3,V5).

% list_147
out(V0,V1,V2):- succ(V3,V2),add(V3,V2,V1),in(V0,V3,V4).
out(V0,V1,V2):- in(V0,V5,V2),add(V5,V1,V3),add(V1,V4,V3),
                add(V4,V5,V1).

% list_158
out(V0,V1,V2):- c0(V2),in(V0,V1,V4),succ(V1,V3),lt(V3,V4).
out(V0,V1,V2):- c1(V2),succ(V1,V3),in(V0,V1,V3).
out(V0,V1,V2):- c0(V2),in(V0,V1,V4),add(V4,V3,V1).

% list_160
out(V0,V1,V2):- in(V0,V1,V3),add(V3,V2,V4),c99(V4).

% list_170
out(V0,V1,V2):- c1(V1),in(V0,V3,V2),end_pos(V0,V4),
                succ(V3,V4).
out(V0,V1,V2):- c0(V1),in(V0,V1,V2).

% list_176
out(V0,V1,V2):- succ(V1,V4),in(V0,V4,V5),add(V3,V5,V2),
                in(V0,V1,V3).

% list_181
out(V0,V1,V2):- in(V0,V1,V2),add(V2,V5,V4),add(V3,V5,V1),
                add(V3,V2,V4).
out(V0,V1,V2):- c3(V1),c7(V3),in(V0,V3,V2).
out(V0,V1,V2):- c3(V3),in(V0,V3,V2),add(V1,V3,V4),
                end_pos(V0,V4).
out(V0,V1,V2):- c1(V3),in(V0,V3,V2),succ(V1,V4),
                end_pos(V0,V4).
out(V0,V1,V2):- c1(V1),in(V0,V3,V2),succ(V3,V4),
                end_pos(V0,V4).
out(V0,V1,V2):- in(V0,V3,V2),c5(V3),add(V1,V3,V4),
                end_pos(V0,V4).

% list_182
out(V0,V1,V2):- in(V0,V3,V2),succ(V1,V4),add(V5,V4,V3),
                add(V4,V1,V5).

% list_187
out(V0,V1,V2):- end_pos(V0,V4),succ(V5,V1),add(V3,V4,V5),
                in(V0,V3,V2).
out(V0,V1,V2):- in(V0,V1,V2).
out(V0,V1,V2):- c0(V2),end_pos(V0,V1).

% list_194
out(V0,V1,V2):- end_pos(V0,V2),c0(V1).
out(V0,V1,V2):- end_pos(V0,V3),in(V0,V4,V2),add(V4,V1,V3).
out(V0,V1,V2):- end_pos(V0,V2),succ(V2,V1).

% list_195
```

```
out(V0,V1,V2):- c4(V1),c99(V2).
out(V0,V1,V2):- c23(V2),c1(V1).
out(V0,V1,V2):- c2(V1),c68(V2).
out(V0,V1,V2):- c3(V1),c42(V2).
out(V0,V1,V2):- c5(V1),c71(V2).
out(V0,V1,V2):- c0(V1),in(V0,V1,V2).
out(V0,V1,V2):- c6(V1),end_pos(V0,V4),succ(V3,V4),
                in(V0,V3,V2).

% list_196
out(V0,V1,V2):- in(V0,V3,V2),c3(V4),add(V3,V4,V1).
out(V0,V1,V2):- c17(V2),c0(V1).
out(V0,V1,V2):- c2(V1),c82(V2).
out(V0,V1,V2):- c1(V1),c38(V2).
out(V0,V1,V2):- c27(V2),end_pos(V0,V3),
                add(V3,V4,V1),c5(V4).
out(V0,V1,V2):- c1(V2),c3(V4),end_pos(V0,V3),
                add(V3,V4,V1).
out(V0,V1,V2):- c55(V2),c4(V4),end_pos(V0,V3),
                add(V3,V4,V1).

% list_212
out(V0,V1,V2):- in(V0,V1,V2),c99(V4),add(V1,V4,V3).
out(V0,V1,V2):- c3(V2),c4(V5),add(V3,V1,V5),add(V4,V3,V1).
out(V0,V1,V2):- in(V0,V4,V2),c4(V3),lt(V3,V1),c3(V5),
                add(V4,V5,V1).

% list_222
out(V0,V1,V2):- end_pos(V0,V2),in(V0,V1,V3).
```

POPPER with our relational representation learns the following programs on the *strings* domain:

```
% string_1
out(V0,V1,V2):- succ(V1,V3),in(V0,V3,V2).

% string_2
out(V0,V1,V2):- in(V0,V4,V2),lt(V1,V4),add(V1,V5,V4),
                in(V0,V5,V3),is_uppercase(V3).

% string_3
out(V0,V1,V2):- c_0(V1),change_case(V3,V2),in(V0,V1,V3).

% string_4
out(V0,V1,V2):- in(V0,V1,V2),succ(V3,V1).
out(V0,V1,V2):- c_0(V1),change_case(V3,V2),in(V0,V1,V3).

% string_5
out(V0,V1,V2):- in(V0,V1,V2),change_case(V2,V3).

% string_6
out(V0,V1,V2):- in(V0,V5,V2),lt(V1,V5),add(V1,V4,V5),
                in(V0,V4,V3),is_uppercase(V3).

% string_9
out(V0,V1,V2):- in(V0,V1,V2),is_space(V5),in(V0,V3,V5),
                in(V0,V4,V5),lt(V4,V3),lt(V1,V4).

% string_23
out(V0,V1,V2):- c_1(V1),end_position(V0,V4),
                succ(V3,V4),in(V0,V3,V2).
out(V0,V1,V2):- c_0(V1),c_2(V3),end_position(V0,V5),
                add(V3,V4,V5),in(V0,V4,V2).

% string_24
out(V0,V1,V2):- c_4(V4),add(V1,V4,V5),in(V0,V5,V2),
                c_8(V3),end_position(V0,V3).
out(V0,V1,V2):- c_3(V4),add(V1,V4,V5),in(V0,V5,V2),
                end_position(V0,V3),c_7(V3).

% string_25
out(V0,V1,V2):- is_number(V2),c_0(V1),in(V0,V3,V2),
                succ(V4,V3),add(V3,V4,V5).

% string_29
out(V0,V1,V2):- c_0(V1),change_case(V4,V2),in(V0,V3,V2).
out(V0,V1,V2):- is_lowercase(V2),in(V0,V3,V2),
                add(V1,V5,V3),in(V0,V5,V4),
                is_uppercase(V4).

% string_30
out(V0,V1,V2):- is_number(V2),in(V0,V1,V2),
                c_7(V3),add(V1,V4,V3).

% string_36
out(V0,V1,V2):- in(V0,V1,V2),c_0(V1).
out(V0,V1,V2):- change_case(V5,V2),in(V0,V3,V2),
                add(V1,V4,V3),succ(V4,V3).

% string_43
out(V0,V1,V2):- in(V0,V1,V2),is_space(V5),in(V0,V3,V5),
                lt(V1,V3),in(V0,V4,V5),lt(V3,V4).

% string_47
out(V0,V1,V2):- is_lowercase(V2),in(V0,V1,V2).
out(V0,V1,V2):- in(V0,V1,V3),change_case(V2,V3).

% string_49
out(V0,V1,V2):- in(V0,V1,V2),is_number(V2),
                c_9(V4),add(V1,V3,V4).
out(V0,V1,V2):- in(V0,V1,V2),is_comma(V2),
                c_5(V4),add(V1,V3,V4).

% string_55
out(V0,V1,V2):- is_lowercase(V2),c_9(V4),add(V1,V4,V3),
                succ(V3,V5),in(V0,V5,V2).
out(V0,V1,V2):- is_fullstop(V2),end_position(V0,V4),
                c_8(V3),add(V1,V3,V5),add(V3,V5,V4).

% string_56
out(V0,V1,V2):- c_4(V3),add(V1,V3,V4),in(V0,V4,V2).

% string_63
out(V0,V1,V2):- c_4(V4),lt(V1,V4),add(V1,V4,V3),
                in(V0,V3,V2).

% string_71
out(V0,V1,V2):- c_7(V4),add(V1,V4,V5),add(V4,V5,V3),
                in(V0,V3,V2).

% string_81
out(V0,V1,V2):- c_0(V1),is_uppercase(V2),in(V0,V5,V2),
                succ(V4,V5),lt(V3,V4).

% string_87
out(V0,V1,V2):- c_8(V4),add(V1,V4,V3),in(V0,V3,V2),
                in(V0,V4,V5),is_uppercase(V5).
out(V0,V1,V2):- is_lowercase(V2),c_9(V4),succ(V4,V5),
                add(V1,V5,V3),lt(V5,V3),in(V0,V3,V2).
out(V0,V1,V2):- c_8(V4),add(V1,V4,V3),in(V0,V3,V2),
                in(V0,V4,V5),is_lowercase(V5).
```

```
out(V0,V1,V2):- c_0(V1),is_fullstop(V5),in(V0,V3,V2),
                succ(V4,V3),in(V0,V4,V5).
out(V0,V1,V2):- c_9(V4),end_position(V0,V4),c_6(V5),
                add(V1,V5,V3),in(V0,V3,V2).
out(V0,V1,V2):- c_2(V1),in(V0,V1,V2),c_8(V4),
                in(V0,V4,V3),is_fullstop(V3).
out(V0,V1,V2):- c_7(V4),end_position(V0,V4),c_5(V5),
                add(V1,V5,V3),in(V0,V3,V2).

% string_91
out(V0,V1,V2):- c_1(V1),c_2(V3),in(V0,V3,V2).
out(V0,V1,V2):- in(V0,V1,V2),c_0(V1).

% string_92
out(V0,V1,V2):- end_position(V0,V3),add(V1,V4,V3),
                succ(V4,V3),in(V0,V4,V2).
out(V0,V1,V2):- in(V0,V1,V2),c_0(V1).

% string_94
out(V0,V1,V2):- in(V0,V3,V2),add(V1,V4,V3),
                add(V4,V5,V1),add(V1,V5,V4).

% string_96
out(V0,V1,V2):- c_2(V1),c_9(V3),add(V1,V3,V4),
                add(V1,V4,V5),in(V0,V5,V2).
out(V0,V1,V2):- c_7(V1),c_4(V3),add(V1,V3,V4),
                add(V1,V4,V5),in(V0,V5,V2).
out(V0,V1,V2):- c_0(V1),c_4(V3),c_7(V4),
                add(V3,V4,V5),in(V0,V5,V2).
out(V0,V1,V2):- c_5(V1),c_9(V3),c_7(V4),
                add(V3,V4,V5),in(V0,V5,V2).
out(V0,V1,V2):- c_6(V1),c_8(V3),c_9(V4),
                add(V3,V4,V5),in(V0,V5,V2).
out(V0,V1,V2):- c_4(V1),c_9(V3),c_6(V4),
                add(V3,V4,V5),in(V0,V5,V2).
out(V0,V1,V2):- c_3(V1),c_8(V3),c_6(V4),
                add(V3,V4,V5),in(V0,V5,V2).
out(V0,V1,V2):- c_8(V1),c_9(V3),in(V0,V5,V2),
                add(V3,V4,V5),lt(V3,V4).
out(V0,V1,V2):- c_1(V1),c_5(V3),c_7(V4),
                add(V3,V4,V5),in(V0,V5,V2).

% string_97
out(V0,V1,V2):- change_case(V2,V4),c_9(V3),
                add(V1,V3,V5),in(V0,V5,V2),lt(V3,V5).
out(V0,V1,V2):- c_0(V1),change_case(V3,V2),
                c_9(V4),in(V0,V4,V3).
out(V0,V1,V2):- is_space(V2),c_7(V1),in(V0,V3,V2).

% string_98
out(V0,V1,V2):- c_0(V1),c_8(V3),in(V0,V3,V2).

% string_99
out(V0,V1,V2):- is_number(V2),in(V0,V3,V2),add(V1,V4,V3),c_0(V4).

% string_102
out(V0,V1,V2):- c_3(V4),add(V1,V4,V3),in(V0,V3,V2).

% string_103
out(V0,V1,V2):- is_number(V2),in(V0,V1,V2),
                c_1(V3),add(V1,V4,V3).

% string_106
out(V0,V1,V2):- is_at(V3),in(V0,V5,V2),
                add(V1,V4,V5),in(V0,V4,V3).

% string_108
out(V0,V1,V2):- in(V0,V5,V2),add(V1,V3,V5),
                add(V3,V4,V1),add(V1,V4,V3).

% string_113
out(V0,V1,V2):- succ(V1,V3),in(V0,V3,V4),
                change_case(V4,V2),lt(V5,V1).
out(V0,V1,V2):- c_2(V5),add(V4,V5,V1),lt(V4,V5),
                add(V1,V4,V3),in(V0,V3,V2).
out(V0,V1,V2):- in(V0,V1,V3),change_case(V3,V2),
                succ(V1,V5),add(V1,V5,V4).

% string_116
out(V0,V1,V2):- is_number(V2),c_8(V4),add(V1,V4,V5),
                add(V4,V5,V3),in(V0,V3,V2).

% string_117
out(V0,V1,V2):- c_0(V1),in(V0,V1,V3),change_case(V3,V2).
out(V0,V1,V2):- in(V0,V1,V2),succ(V4,V1),in(V0,V4,V3),
                change_case(V3,V5).
out(V0,V1,V2):- succ(V3,V1),change_case(V5,V2),
                in(V0,V1,V5),is_space(V4),in(V0,V3,V4).

% string_120
out(V0,V1,V2):- c_1(V1),c_4(V5),end_position(V0,V3),
                add(V4,V5,V3),in(V0,V4,V2).
out(V0,V1,V2):- c_2(V1),end_position(V0,V5),add(V1,V3,V5),
                succ(V4,V3),in(V0,V4,V2).
out(V0,V1,V2):- c_3(V1),end_position(V0,V5),add(V1,V3,V5),
                succ(V3,V4),in(V0,V4,V2).
out(V0,V1,V2):- c_0(V1),end_position(V0,V3),c_5(V5),
                add(V4,V5,V3),in(V0,V4,V2).

% string_121
out(V0,V1,V2):- in(V0,V1,V2),is_lowercase(V3),
                in(V0,V5,V3),add(V1,V4,V5).

% string_122
out(V0,V1,V2):- in(V0,V1,V2),is_space(V4),
                in(V0,V3,V4),lt(V1,V3).

% string_125
out(V0,V1,V2):- c_7(V1),is_lowercase(V2),c_9(V3),
                in(V0,V3,V5),is_lowercase(V5),
                succ(V3,V4),in(V0,V4,V2).
out(V0,V1,V2):- is_lowercase(V2),c_7(V4),lt(V1,V4),
                c_3(V3),add(V1,V3,V5),in(V0,V5,V2).
out(V0,V1,V2):- c_0(V1),c_3(V3),in(V0,V3,V2).

% string_128
out(V0,V1,V2):- c_6(V4),add(V1,V3,V4),
                add(V1,V4,V5),in(V0,V5,V2).
out(V0,V1,V2):- is_number(V2),c_6(V4),add(V1,V4,V5),
                in(V0,V5,V2),add(V1,V5,V3).

% string_132
out(V0,V1,V2):- is_number(V2),succ(V1,V4),in(V0,V4,V2),
                in(V0,V1,V3),in(V0,V5,V3),lt(V4,V5).
out(V0,V1,V2):- in(V0,V1,V2),is_comma(V4),c_4(V5),
                in(V0,V3,V4),lt(V3,V5),lt(V1,V3).
out(V0,V1,V2):- is_number(V2),c_7(V4),lt(V1,V4),
                in(V0,V5,V2),lt(V1,V5),add(V1,V3,V5),
                lt(V3,V1).
```

% string_136
out(V0,V1,V2):- c_0(V1),is_fullstop(V4),in(V0,V3,V2),
                succ(V5,V3),in(V0,V5,V4).
out(V0,V1,V2):- end_position(V0,V5),succ(V4,V5),
                add(V1,V3,V4),in(V0,V3,V2),succ(V3,V4).
out(V0,V1,V2):- c_2(V1),end_position(V0,V4),
                succ(V3,V4),in(V0,V3,V2).

% string_137
out(V0,V1,V2):- in(V0,V1,V2),change_case(V4,V5),
                in(V0,V3,V5),lt(V1,V3).

% string_139
out(V0,V1,V2):- in(V0,V1,V2),c_8(V4),add(V1,V3,V4).
out(V0,V1,V2):- in(V0,V1,V2),c_9(V1),change_case(V2,V3).

% string_144
out(V0,V1,V2):- c_2(V4),change_case(V2,V5),
                add(V1,V4,V3),in(V0,V3,V2).
out(V0,V1,V2):- succ(V1,V4),c_9(V3),lt(V1,V3),
                succ(V4,V5),in(V0,V5,V2).
out(V0,V1,V2):- c_9(V1),c_2(V4),add(V1,V4,V3),
                in(V0,V3,V2).

% string_149
out(V0,V1,V2):- is_fullstop(V2),c_6(V4),add(V1,V4,V3),
                in(V0,V3,V2).
out(V0,V1,V2):- is_number(V2),c_6(V4),add(V1,V4,V3),
                in(V0,V3,V2).

% string_151
out(V0,V1,V2):- in(V0,V3,V2),add(V1,V4,V3),
                add(V4,V5,V1),add(V1,V5,V4).

% string_152
out(V0,V1,V2):- c_8(V4),add(V1,V4,V5),
                add(V4,V5,V3),in(V0,V3,V2).

% string_154
out(V0,V1,V2):- in(V0,V1,V2),is_lowercase(V5),
                in(V0,V3,V5),add(V1,V4,V3).

% string_156
out(V0,V1,V2):- c_7(V1),in(V0,V3,V2),add(V1,V4,V3),
                in(V0,V4,V5),is_uppercase(V5).
out(V0,V1,V2):- c_6(V1),in(V0,V3,V2),add(V1,V4,V3),
                in(V0,V4,V5),is_uppercase(V5).
out(V0,V1,V2):- c_3(V1),c_7(V4),end_position(V0,V5),
                add(V3,V4,V5),in(V0,V3,V2).
out(V0,V1,V2):- c_0(V1),is_uppercase(V2),c_8(V5),
                in(V0,V3,V2),add(V3,V5,V4).
out(V0,V1,V2):- c_5(V1),end_position(V0,V4),
                add(V1,V3,V4),in(V0,V3,V2).
out(V0,V1,V2):- c_1(V1),in(V0,V3,V2),add(V1,V4,V3),
                in(V0,V4,V5),is_uppercase(V5).
out(V0,V1,V2):- c_8(V1),in(V0,V3,V2),add(V1,V4,V3),
                in(V0,V4,V5),is_uppercase(V5).
out(V0,V1,V2):- c_2(V1),end_position(V0,V5),c_8(V4),
                add(V3,V4,V5),in(V0,V3,V2).
out(V0,V1,V2):- c_9(V1),end_position(V0,V4),
                succ(V3,V4),in(V0,V3,V2).
out(V0,V1,V2):- c_4(V1),in(V0,V3,V2),add(V1,V4,V3),
                in(V0,V4,V5),is_uppercase(V5).

% string_162
out(V0,V1,V2):- is_lowercase(V2),end_position(V0,V3),
                c_3(V5),succ(V5,V1),add(V4,V5,V3),
                in(V0,V4,V2).
out(V0,V1,V2):- c_4(V1),c_7(V3),add(V1,V3,V5),
                in(V0,V5,V2),in(V0,V4,V2),lt(V4,V1).
out(V0,V1,V2):- is_lowercase(V2),c_7(V5),add(V1,V5,V4),
                in(V0,V4,V2),add(V1,V3,V5),lt(V1,V3).
out(V0,V1,V2):- c_9(V3),c_7(V4),add(V1,V4,V5),
                in(V0,V5,V2),lt(V5,V3).

% string_163
out(V0,V1,V2):- is_uppercase(V2),c_7(V5),add(V1,V5,V4),
                add(V4,V5,V3),in(V0,V3,V2).
out(V0,V1,V2):- c_7(V3),lt(V1,V3),add(V1,V3,V5),
                in(V0,V5,V2),in(V0,V4,V2),lt(V5,V4).
out(V0,V1,V2):- c_4(V1),c_7(V3),add(V1,V3,V4),
                in(V0,V4,V2),in(V0,V5,V2),lt(V5,V1).
out(V0,V1,V2):- c_7(V5),succ(V1,V5),add(V1,V5,V4),
                add(V4,V5,V3),in(V0,V3,V2).
out(V0,V1,V2):- is_number(V2),c_7(V5),add(V1,V5,V4),
                add(V4,V5,V3),in(V0,V3,V2).
out(V0,V1,V2):- c_7(V4),add(V1,V5,V4),lt(V1,V5),
                add(V1,V4,V3),in(V0,V3,V2).
out(V0,V1,V2):- c_7(V4),lt(V4,V1),add(V1,V4,V5),
                add(V4,V5,V3),in(V0,V3,V2).

% string_168
out(V0,V1,V2):- in(V0,V1,V2),c_5(V5),add(V1,V5,V3),
                in(V0,V3,V4),is_number(V4).
out(V0,V1,V2):- in(V0,V1,V2),is_lowercase(V2),
                add(V1,V4,V3),add(V1,V3,V5).
out(V0,V1,V2):- in(V0,V1,V2),is_uppercase(V4),
                in(V0,V3,V4),add(V1,V5,V3),lt(V1,V5).

% string_173
out(V0,V1,V2):- c_6(V1),end_position(V0,V4),
                succ(V3,V4),in(V0,V3,V2).
out(V0,V1,V2):- c_5(V1),end_position(V0,V4),c_2(V3),
                add(V3,V5,V4),in(V0,V5,V2).
out(V0,V1,V2):- c_2(V1),end_position(V0,V4),c_5(V3),
                add(V3,V5,V4),in(V0,V5,V2).
out(V0,V1,V2):- c_3(V1),c_4(V4),end_position(V0,V3),
                add(V4,V5,V3),in(V0,V5,V2).
out(V0,V1,V2):- succ(V4,V1),c_3(V4),end_position(V0,V3),
                add(V4,V5,V3),in(V0,V5,V2).
out(V0,V1,V2):- c_0(V1),c_7(V4),end_position(V0,V3),
                add(V4,V5,V3),in(V0,V5,V2).
out(V0,V1,V2):- c_1(V1),c_6(V4),end_position(V0,V3),
                add(V4,V5,V3),in(V0,V5,V2).

% string_177
out(V0,V1,V2):- in(V0,V1,V2),is_comma(V5),in(V0,V3,V5),
                in(V0,V4,V5),lt(V1,V4),lt(V4,V3).

% string_178
out(V0,V1,V2):- c_2(V1),c_3(V4),end_position(V0,V3),
                add(V4,V5,V3),in(V0,V5,V2).
out(V0,V1,V2):- c_3(V1),end_position(V0,V3),
                succ(V4,V1),add(V4,V5,V3),in(V0,V5,V2).
out(V0,V1,V2):- c_4(V1),end_position(V0,V4),
                succ(V3,V4),in(V0,V3,V2).
out(V0,V1,V2):- c_0(V1),c_5(V4),end_position(V0,V3),
                add(V4,V5,V3),in(V0,V5,V2).
out(V0,V1,V2):- c_1(V1),end_position(V0,V3),c_4(V4),
                add(V4,V5,V3),in(V0,V5,V2).

```
% string_185
out(V0,V1,V2):- is_number(V2),in(V0,V1,V2),
                end_position(V0,V3),add(V1,V4,V3),
                add(V1,V5,V4).
out(V0,V1,V2):- in(V0,V1,V2),is_comma(V5),
                in(V0,V4,V5),add(V1,V3,V4).

% string_186
out(V0,V1,V2):- c_1(V1),end_position(V0,V4),
                c_4(V3),add(V3,V5,V4),in(V0,V5,V2).
out(V0,V1,V2):- c_0(V1),end_position(V0,V4),c_5(V3),
                add(V3,V5,V4),in(V0,V5,V2).
out(V0,V1,V2):- c_3(V1),end_position(V0,V3),
                add(V1,V4,V3),succ(V4,V5),in(V0,V5,V2).
out(V0,V1,V2):- c_2(V1),end_position(V0,V3),
                add(V1,V4,V3),succ(V5,V4),in(V0,V5,V2).

% string_188
out(V0,V1,V2):- in(V0,V1,V2),c_3(V4),add(V1,V3,V4).

% string_196
out(V0,V1,V2):- in(V0,V1,V2),c_8(V3),add(V1,V4,V3).

% string_206
out(V0,V1,V2):- c_6(V1),c_4(V3),add(V1,V3,V5),
                in(V0,V5,V2),in(V0,V4,V2),lt(V4,V3).
out(V0,V1,V2):- c_7(V1),c_4(V3),add(V1,V3,V5),
                in(V0,V5,V2),in(V0,V4,V2),lt(V4,V3).
out(V0,V1,V2):- is_lowercase(V2),c_6(V4),lt(V1,V4),
                add(V1,V4,V5),in(V0,V5,V2),in(V0,V4,V3),
                is_uppercase(V3).
out(V0,V1,V2):- c_5(V1),c_9(V5),in(V0,V5,V2),
                succ(V4,V1),in(V0,V4,V3),is_uppercase(V3).
out(V0,V1,V2):- is_lowercase(V2),c_5(V4),lt(V1,V4),
                add(V1,V4,V5),in(V0,V5,V2),in(V0,V4,V3),
                is_uppercase(V3).
out(V0,V1,V2):- c_9(V4),lt(V1,V4),add(V1,V4,V5),
                in(V0,V5,V2),in(V0,V4,V3),
                is_uppercase(V3).
out(V0,V1,V2):- is_lowercase(V2),c_7(V4),lt(V1,V4),
                add(V1,V4,V5),in(V0,V5,V2),in(V0,V4,V3),
                is_uppercase(V3).
out(V0,V1,V2):- c_4(V1),c_2(V4),in(V0,V4,V2),
                in(V0,V1,V3),is_uppercase(V3).
out(V0,V1,V2):- c_0(V1),is_uppercase(V2),c_9(V3),
                in(V0,V5,V2),succ(V4,V5),lt(V4,V3).
out(V0,V1,V2):- c_4(V4),lt(V1,V4),add(V1,V4,V5),
                in(V0,V5,V2),in(V0,V4,V3),
                is_uppercase(V3).

% string_215
out(V0,V1,V2):- c_0(V1),c_4(V3),end_position(V0,V4),
                add(V3,V5,V4),in(V0,V5,V2).
out(V0,V1,V2):- c_2(V1),end_position(V0,V4),
                add(V1,V3,V4),in(V0,V3,V2).
out(V0,V1,V2):- c_1(V1),c_3(V4),end_position(V0,V3),
                add(V4,V5,V3),in(V0,V5,V2).

% string_222
out(V0,V1,V2):- c_0(V1),change_case(V4,V2),
                c_9(V3),in(V0,V3,V4).
out(V0,V1,V2):- is_lowercase(V2),c_9(V5),
                add(V1,V5,V3),lt(V5,V3),in(V0,V3,V2),
                c_7(V4),lt(V1,V4).

% string_226
out(V0,V1,V2):- c_0(V1),change_case(V4,V2),in(V0,V3,V2).
out(V0,V1,V2):- c_9(V5),add(V1,V5,V3),lt(V5,V3),
                in(V0,V3,V2),change_case(V2,V4).
out(V0,V1,V2):- c_0(V1),c_9(V3),change_case(V4,V2),i
                n(V0,V3,V4).
out(V0,V1,V2):- is_space(V2),c_7(V1),in(V0,V3,V2).

% string_229
out(V0,V1,V2):- in(V0,V1,V2),is_space(V5),
                in(V0,V3,V5),in(V0,V4,V5),lt(V4,V3),
                lt(V1,V4).

% string_237
out(V0,V1,V2):- is_lowercase(V2),in(V0,V1,V2),
                is_lowercase(V4),in(V0,V5,V4),
                add(V1,V3,V5),lt(V1,V3).
out(V0,V1,V2):- in(V0,V1,V2),c_9(V3),add(V1,V4,V3).

% string_239
out(V0,V1,V2):- in(V0,V1,V2),c_3(V4),add(V1,V3,V4).
out(V0,V1,V2):- c_4(V5),add(V3,V5,V1),
                succ(V1,V4),in(V0,V4,V2).

% string_242
out(V0,V1,V2):- in(V0,V1,V2).

% string_246
out(V0,V1,V2):- in(V0,V5,V2),add(V1,V3,V5),
                add(V3,V4,V1),add(V1,V4,V3).

% string_252
out(V0,V1,V2):- c_5(V1),end_position(V0,V3),
                c_2(V4),add(V4,V5,V3),in(V0,V5,V2).
out(V0,V1,V2):- c_0(V1),is_space(V4),in(V0,V5,V2),
                succ(V3,V5),in(V0,V3,V4).
out(V0,V1,V2):- c_6(V1),end_position(V0,V4),
                succ(V3,V4),in(V0,V3,V2).
out(V0,V1,V2):- c_3(V1),end_position(V0,V4),
                add(V1,V3,V4),succ(V5,V3),in(V0,V5,V2).
out(V0,V1,V2):- c_4(V1),end_position(V0,V4),
                add(V1,V3,V4),succ(V3,V5),in(V0,V5,V2).
out(V0,V1,V2):- c_2(V1),c_5(V4),end_position(V0,V3),
                add(V4,V5,V3),in(V0,V5,V2).
out(V0,V1,V2):- c_1(V1),c_6(V4),end_position(V0,V3),
                add(V4,V5,V3),in(V0,V5,V2).

% string_295
out(V0,V1,V2):- c_8(V3),add(V1,V3,V4),in(V0,V4,V2),
                in(V0,V3,V5),is_lowercase(V5).
out(V0,V1,V2):- c_2(V3),add(V1,V3,V4),in(V0,V4,V2),
                in(V0,V3,V5),is_lowercase(V5).
out(V0,V1,V2):- c_5(V1),change_case(V5,V2),in(V0,V3,V5),
                add(V1,V3,V4),in(V0,V4,V2).
out(V0,V1,V2):- in(V0,V1,V2),c_0(V3),in(V0,V3,V5),
                change_case(V5,V4).
out(V0,V1,V2):- c_1(V1),c_7(V4),add(V1,V3,V4),
                in(V0,V3,V2),add(V3,V4,V5),in(V0,V5,V2).
out(V0,V1,V2):- end_position(V0,V4),add(V1,V3,V4),
                in(V0,V3,V2),lt(V1,V3),succ(V5,V3),
                in(V0,V5,V2).
out(V0,V1,V2):- c_6(V3),add(V1,V3,V4),in(V0,V4,V2),
                in(V0,V3,V5),is_lowercase(V5).
out(V0,V1,V2):- c_0(V1),is_lowercase(V2),in(V0,V1,V5),
```

```
                in(V0,V3,V2),in(V0,V4,V5),lt(V1,V4).

% string_300
out(V0,V1,V2):- in(V0,V1,V2),c_8(V4),add(V1,V4,V5),i
                n(V0,V5,V3),in(V0,V4,V3).
out(V0,V1,V2):- in(V0,V1,V2),c_9(V4),lt(V1,V4),
                in(V0,V4,V3),is_number(V3).
out(V0,V1,V2):- in(V0,V1,V2),is_lowercase(V2),c_9(V4),
                lt(V4,V1),is_number(V3),in(V0,V5,V3),
                lt(V1,V5).
out(V0,V1,V2):- in(V0,V1,V2),c_4(V4),
                add(V1,V3,V4),change_case(V5,V2).
out(V0,V1,V2):- in(V0,V1,V2),is_uppercase(V2),c_9(V4),
                add(V3,V4,V1),add(V1,V3,V5),
                end_position(V0,V5).
out(V0,V1,V2):- in(V0,V1,V2),is_lowercase(V2),
                is_lowercase(V3),in(V0,V5,V3),
                in(V0,V4,V3),lt(V4,V5),lt(V1,V4).
out(V0,V1,V2):- in(V0,V1,V2),c_8(V4),change_case(V3,V2),
                end_position(V0,V5),add(V1,V4,V5).
out(V0,V1,V2):- in(V0,V1,V2),is_uppercase(V2),
                is_lowercase(V3),in(V0,V5,V3),
                in(V0,V4,V3),lt(V4,V5),lt(V1,V4).

% string_323
out(V0,V1,V2):- in(V0,V1,V2),is_number(V3),
                in(V0,V4,V3),lt(V1,V4).

% string_326
out(V0,V1,V2):- is_number(V2),in(V0,V1,V2),
                c_6(V3),lt(V1,V3).

% string_327
out(V0,V1,V2):- in(V0,V1,V2),change_case(V5,V2),
                c_6(V4),add(V1,V3,V4).
out(V0,V1,V2):- in(V0,V1,V2),change_case(V2,V5),
                c_6(V4),add(V1,V3,V4).
```